\documentclass[11pt]{article}

\usepackage[margin=1in]{geometry}
\usepackage{amsmath,amssymb,amsfonts,amsthm}
\usepackage{graphicx}
\usepackage{booktabs}
\usepackage{hyperref}
\usepackage{cleveref}
\usepackage{caption}
\usepackage{subcaption}
\usepackage{xcolor}
\usepackage{authblk}
\usepackage{algorithm}
\usepackage{algorithmic}
\usepackage{multirow}
\usepackage{natbib}
\newcommand{\R}{\mathbb{R}}

\newcommand{\norm}[1]{\lVert #1 \rVert}
\newcommand{\comm}[2]{[#1, #2]}
\newcommand{\defect}{\mathcal{D}}
\newcommand{\grokstep}{t_{\text{grok}}}
\newcommand{\onsetstep}{t_{\text{onset}}}
\newcommand{\leadtime}{\Delta t}

\newtheorem{definition}{Definition}

\title{Early-Warning Signals of Grokking via Loss-Landscape Geometry}

\author{Yongzhong Xu\thanks{abbyxu@gmail.com}}
\date{}

\begin{document}
\maketitle

\begin{abstract}
Grokking refers to the phenomenon in which neural networks abruptly transition from memorization to generalization after extended training. Recent work on modular arithmetic has linked grokking to prolonged confinement on low-dimensional execution manifolds in weight space, followed by a sudden escape into a generalizing solution. A central open question is whether this geometric mechanism extends beyond arithmetic tasks.

We investigate this question on two structurally distinct sequence-learning problems: the SCAN compositional generalization benchmark and Dyck-1 language depth prediction. Across both tasks and a wide range of learning rates, we find that the commutator defect---a measure of loss-landscape curvature derived from the non-commutativity of successive gradient updates---reliably rises well before the onset of generalization. The lead time between defect onset and grokking follows a power-law relationship with the grokking timescale, with exponents $\alpha \approx 1.18$ for SCAN ($R^2 = 0.990$, $n = 11$) and $\alpha \approx 1.13$ for Dyck ($R^2 = 0.908$, $n = 14$). Combined with prior results on modular arithmetic, all three task families exhibit superlinear scaling ($\alpha > 1$), yielding advance warning windows of 90--97\% at slow learning rates.

Weight-space principal component analysis reveals task-dependent spectral dynamics: PC1 de-concentration precedes grokking in Dyck and modular arithmetic but follows it in SCAN, indicating that spectral concentration is not a universal precursor. In contrast, the commutator defect consistently anticipates generalization across all settings. A three-basis integrability decomposition shows that defect spikes reflect structured non-commutativity within the learning subspace, characteristic of the grokking regime.

Finally, causal interventions on both SCAN and Dyck demonstrate a mechanistic role for the defect: perturbations that amplify non-commutativity accelerate grokking (by ${\sim}32\%$ on SCAN and ${\sim}50\%$ on Dyck), whereas suppressing orthogonal gradient flow delays or prevents generalization. The three task families form a spectrum of causal sensitivity: modular arithmetic is rigid (boosting has no effect), Dyck is responsive (both mild and aggressive boosting accelerate), and SCAN is intermediate (mild boosting accelerates but aggressive boosting destabilizes). Across all tasks, suppression delays or prevents grokking, establishing necessity as a universal finding. Together with prior results, our findings establish the commutator defect as a robust, architecture-agnostic, and causally implicated early-warning signal for grokking in transformer-based models.
\end{abstract}

\section{Introduction}
\label{sec:intro}

Grokking---the phenomenon where neural networks trained on small datasets first memorize the training set and then, long after achieving perfect training accuracy, suddenly generalize to the test set---was first reported by~\citet{power2022grokking} in modular arithmetic tasks. The phenomenon challenges conventional accounts of generalization because standard training metrics (loss, accuracy) provide no advance warning: a model with perfect training accuracy and poor test accuracy might be on the verge of grokking, or might never generalize at all.

In our prior work~\citep{xu2026integrability}, we proposed a geometric account of grokking centered on \emph{transverse curvature dynamics}. Working on modular arithmetic (six binary operations mod 97), we showed that:
\begin{enumerate}
  \item Weight-space trajectories during grokking lie on a rank-1 \emph{execution manifold} (PC1 captures 68--83\% of variance);
  \item This manifold exhibits empirical invariance: commutator defect vectors---measuring the non-commutativity of successive gradient updates---are predominantly orthogonal to it ($\rho \approx 1.000$);
  \item Curvature explodes in the normal bundle (10--1000$\times$), with onset preceding generalization by 600--1600 steps;
  \item The lead time obeys a power law $\leadtime \propto \grokstep^\alpha$ with $\alpha = 1.27 \pm 0.03$ ($R^2 = 0.97$, $n = 43$) across a 300$\times$ learning-rate sweep;
  \item Causal interventions confirm that orthogonal gradient flow is necessary (suppression prevents grokking) but curvature boosting alone is not sufficient (no acceleration).
\end{enumerate}
A fundamental open question raised by that work is whether these findings are specific to modular arithmetic and encoder-only transformers, or whether they reflect a universal geometric mechanism of grokking. The original paper noted that ``preliminary experiments on Dyck languages and the SCAN compositional-generalization benchmark suggest that qualitatively similar low-dimensional confinement and transverse curvature dynamics arise beyond modular arithmetic; a systematic investigation of these settings is ongoing.''

This paper provides that systematic investigation. We extend the commutator defect analysis to two structurally distinct tasks:
\begin{enumerate}
  \item \textbf{SCAN}~\citep{lake2018generalization}: A compositional generalization benchmark mapping natural language commands (e.g., ``jump twice'') to action sequences (e.g., \texttt{JUMP JUMP}), trained with an encoder--decoder transformer.
  \item \textbf{Dyck-1 depth prediction}: A formal language task where a causal (decoder-only) transformer must predict the current stack depth at each position in a parenthesis sequence, trained on only 50 examples.
\end{enumerate}
These tasks differ from modular arithmetic in architecture (encoder--decoder and causal vs.\ encoder-only), input domain (natural language and formal language vs.\ numerical), output type (sequence generation and per-token classification vs.\ single-token classification), and dataset structure.

\paragraph{Key contributions.} Our work makes six main contributions:
\begin{enumerate}
  \item \textbf{Universality of defect onset.} On both SCAN and Dyck, the commutator defect onset reliably precedes the generalization transition across all tested learning rates---15 of 15 grokking runs on Dyck and 11 of 11 on SCAN show positive lead time.

  \item \textbf{Super-linear scaling.} The lead time obeys a power law with super-linear exponent on both tasks: $\alpha \approx 1.18$ ($R^2 = 0.990$) for SCAN and $\alpha \approx 1.13$ ($R^2 = 0.908$) for Dyck. Combined with modular arithmetic ($\alpha \approx 1.3$), all three task families exhibit $\alpha > 1$.

  \item \textbf{PC1 dissociation.} Weight-space PC1 de-concentration precedes grokking on Dyck (as in modular arithmetic) but \emph{follows} grokking on SCAN, demonstrating that the commutator defect is a more universal signal than spectral concentration.

  \item \textbf{Integrability decomposition.} A three-basis analysis (Weight SVD, $\Delta W$ SVD, Gradient SVD) shows that the defect spike reflects structured non-commutativity aligned with the learning subspace (exec/random ratio 2--3$\times$), specific to the grokking regime.

  \item \textbf{Causal evidence.} On both SCAN and Dyck, perturbations that amplify the commutator accelerate grokking (${\sim}32\%$ on SCAN, ${\sim}50\%$ on Dyck); perturbations that suppress orthogonal gradient flow delay or prevent it. The three task families (including modular arithmetic) form a spectrum of causal sensitivity, while universally confirming necessity.

  \item \textbf{Instability regime.} At high learning rates (Dyck, $\eta = 10^{-2}$), accuracy oscillates rather than transitioning cleanly, motivating sustained-accuracy grokking criteria.
\end{enumerate}

\paragraph{Paper outline.} Section~\ref{sec:background} reviews the geometric framework of~\citet{xu2026integrability}. Section~\ref{sec:setup} describes the experimental setup. Section~\ref{sec:methods} presents the methods: PCA eigenanalysis, commutator defect measurement, and integrability decomposition. Section~\ref{sec:results} presents results in four stages: SCAN (Section~\ref{sec:scan_results}), Dyck (Section~\ref{sec:dyck_results}), geometric analysis (Section~\ref{sec:geometry}), and causal interventions (Section~\ref{sec:interventions}). Section~\ref{sec:comparison} synthesizes the cross-dataset comparison. Section~\ref{sec:spectral} presents a spectral geometry analysis of the attention weight matrices, testing whether spectral symmetry-breaking in $W_Q$ and $W_K$ accompanies grokking. Section~\ref{sec:discussion} discusses implications and connections to broader themes.

\section{Background: Geometric Framework for Grokking}
\label{sec:background}

We review the geometric framework introduced in our prior work~\citep{xu2026integrability} for studying grokking through the lens of loss-landscape curvature. This section defines the key mathematical objects and establishes the theoretical context for the present work.

\subsection{Grokking}

\citet{power2022grokking} first reported grokking on modular arithmetic tasks, showing that small transformers can memorize training data in hundreds of steps but require thousands to generalize. Subsequent work has shown grokking in diverse settings including group operations~\citep{nanda2023progress}, boolean functions, sparse parities~\citep{barak2022hidden}, and compositional generalization~\citep{murty2023grokking}. The phenomenon is sensitive to regularization strength~\citep{liu2022omnigrok}, dataset size~\citep{liu2023grokking}, and learning rate.

\subsection{The Execution Manifold}
\label{sec:execution_manifold}

Following our earlier work~\citep{xu2026integrability}, we define the \emph{execution manifold} as the low-dimensional subspace of parameter space traced by the weight trajectory during training. Formally, given $T$ weight snapshots $\{W_t\}_{t=1}^{T}$ of a weight matrix $W \in \R^{d \times d}$, we form the trajectory matrix of displacements from initialization:
\begin{equation}
  X = \begin{pmatrix} \text{vec}(W_1 - W_0) \\ \vdots \\ \text{vec}(W_T - W_0) \end{pmatrix} \in \R^{T \times d^2},
  \label{eq:trajectory_matrix}
\end{equation}
after column-centering. The SVD $X = U \Sigma V^\top$ yields principal directions $\{v_k\}$. The explained variance ratio of the $k$-th component is $\sigma_k^2 / \sum_i \sigma_i^2$, and $\text{PC1\%} = 100 \times \sigma_1^2 / \sum_i \sigma_i^2$ measures the fraction of trajectory variance captured by a single direction.

\begin{definition}[Execution manifold]
Let $V_K = [v_1, \ldots, v_K]$ denote the top-$K$ right singular vectors of $X$. The \emph{execution manifold} is the affine subspace $\mathcal{M} = \{W_0 + V_K \alpha : \alpha \in \R^K\}$ through the initial weights $W_0$. When $\text{PC1\%}$ is high ($> 70\%$), $\mathcal{M}$ is effectively rank-1: the weight trajectory is confined to a one-dimensional curve in parameter space.
\end{definition}

In modular arithmetic, we found $\text{PC1\%} = 68$--$83\%$ across all grokking conditions~\citep{xu2026integrability}, with z-scores 5--20$\sigma$ above a random-walk null model.

\subsection{The Commutator Defect}
\label{sec:commutator_defect_def}

The commutator defect quantifies the non-commutativity of gradient updates from different mini-batches, providing a local probe of loss-landscape curvature that requires no Hessian computation.

\begin{definition}[Commutator defect]
\label{def:commutator}
Given parameters $\theta_0$, loss function $\mathcal{L}$, step size $\eta$, and two independent mini-batches $A, B$, define:
\begin{align}
  \theta_{AB} &= \theta_0 - \eta\, g_A(\theta_0) - \eta\, g_B(\theta_0 - \eta\, g_A(\theta_0)), \label{eq:theta_AB}\\
  \theta_{BA} &= \theta_0 - \eta\, g_B(\theta_0) - \eta\, g_A(\theta_0 - \eta\, g_B(\theta_0)), \label{eq:theta_BA}
\end{align}
where $g_A(\theta) = \nabla_\theta \mathcal{L}_A(\theta)$ is the gradient on mini-batch $A$ at parameters $\theta$. The \emph{scale-normalized commutator defect} is:
\begin{equation}
  \defect(\theta_0; A, B) = \frac{\norm{\theta_{AB} - \theta_{BA}}}{\norm{\eta\, g_A} \cdot \norm{\eta\, g_B}}.
  \label{eq:defect}
\end{equation}
\end{definition}

\paragraph{Geometric interpretation.} To first order in $\eta$, the commutator vector $\delta = \theta_{AB} - \theta_{BA}$ satisfies:
\begin{equation}
  \delta \approx \eta^2 (\nabla g_B \cdot g_A - \nabla g_A \cdot g_B),
  \label{eq:lie_bracket}
\end{equation}
which is the Lie bracket of the gradient vector fields~\citep{xu2026integrability}. $\defect$ is thus proportional to the Lie bracket of stochastic gradient vector fields and serves as a proxy for local nonlinearity of the loss landscape: if the landscape is locally flat (integrable), gradient steps commute and $\defect = 0$. A large defect signals that the order of gradient updates matters---the loss landscape has developed curvature structure that breaks path-independence.

\subsection{Manifold Projection and Invariance}
\label{sec:invariance_def}

To determine whether curvature lives inside or outside the execution manifold, we decompose the commutator vector $\delta$ relative to the PCA basis~\citep{xu2026integrability}.

\begin{definition}[Invariance measure]
Let $B \in \R^{P \times K}$ be an orthonormal basis for the PCA subspace embedded in the full parameter space ($P$ total parameters). Given commutator vector $\delta = \theta_{AB} - \theta_{BA}$, decompose:
\begin{equation}
  \delta = \underbrace{B B^\top \delta}_{\delta_\parallel\ (\text{projected})} + \underbrace{(\delta - B B^\top \delta)}_{\delta_\perp\ (\text{residual})}.
  \label{eq:projection}
\end{equation}
The \emph{invariance measure} is the residual fraction:
\begin{equation}
  \rho = \frac{\norm{\delta_\perp}}{\norm{\delta}}.
  \label{eq:invariance}
\end{equation}
\end{definition}

If $\rho \approx 1$, the commutator is orthogonal to the execution manifold---curvature is confined to the \emph{normal bundle}---and the manifold exhibits \emph{empirical invariance} under the optimization dynamics. If $\rho \approx 0$, curvature lies within the learned subspace.

\begin{definition}[Transverse decoupling]
The optimization dynamics are \emph{transversely decoupled} on the execution manifold $\mathcal{M}$ if commutator defect vectors are confined to its normal bundle ($\rho \approx 1$). In this regime, curvature-induced perturbations act almost entirely orthogonally to $\mathcal{M}$, so that the observed trajectory remains confined to the learned subspace over training.
\end{definition}

In modular arithmetic, we found $\rho \approx 1.000$ within numerical precision across all 36 conditions~\citep{xu2026integrability} (6 operations $\times$ 2 weight-decay settings $\times$ 3 seeds), with the exec/random projection ratio at 1.8--2.9$\times$, confirming that the near-zero parallel component is geometrically structured rather than a dimensionality artifact.

\subsection{Random Subspace Control}

Because any $K$-dimensional subspace of $\R^P$ captures $\sim \sqrt{K/P}$ of a random vector, near-unity $\rho$ alone does not confirm geometric structure. Following~\citet{xu2026integrability}, we compare the PCA-basis projection against random baselines: for each commutator vector $\delta$, we compute the projection fraction onto $N_{\text{rand}} = 5$ random $K$-dimensional orthonormal bases and report the \emph{exec/random ratio} $\text{proj}_{\text{exec}} / \text{proj}_{\text{rand}}$. Values significantly above 1.0 confirm that the PCA subspace captures more commutator energy than expected by chance.

\subsection{Scaling Law}
\label{sec:scaling_def}

We established~\citep{xu2026integrability} that the lead time between defect onset and grokking obeys a power law:
\begin{equation}
  \leadtime = C \cdot \grokstep^\alpha,
  \label{eq:scaling_law}
\end{equation}
where $\leadtime = \grokstep - \onsetstep$ is the lead time, $\grokstep$ is the grokking step, and $\onsetstep$ is the defect onset step. The \emph{lead fraction} $\leadtime / \grokstep$ expresses lead time as a proportion of total training time. In modular arithmetic, $\alpha = 1.27 \pm 0.03$ ($R^2 = 0.97$, $n = 43$) across a 300$\times$ learning-rate sweep: because $\alpha > 1$, the predictive window grows super-linearly with grokking timescale.

\section{Experimental Setup}
\label{sec:setup}

\subsection{Tasks and Datasets}

\paragraph{SCAN Compositional Generalization.}
We use the SCAN simple split~\citep{lake2018generalization}, randomly sampling $N_{\text{train}} = 2{,}048$ command--action pairs from the full dataset of approximately 14,000 examples. The remaining examples form the test set (${\sim}9{,}000$). The command vocabulary contains 21 tokens and the action vocabulary 7 tokens. Maximum command length is 8 tokens and maximum action length is 13 tokens (both plus boundary tokens). The data split is fixed across all runs (data seed = 0).

\paragraph{Dyck-1 Depth Prediction.}
We generate Dyck-1 sequences of length 24 (maximum nesting depth 12) with a fixed random seed. The model must predict the current stack depth at each parenthesis position: a 13-class classification problem (depths 0--12). The vocabulary contains 3 tokens: open parenthesis, close parenthesis, and padding. We use an extreme data-scarcity regime: $N_{\text{train}} = 50$ sequences, $N_{\text{test}} = 5{,}000$ sequences. This large train--test ratio amplifies the grokking effect.

\subsection{Model Architectures}

\paragraph{SCAN.}
We use a standard encoder--decoder transformer with $d_{\text{model}} = 256$, $n_{\text{layers}} = 3$ (each for encoder and decoder), $n_{\text{heads}} = 4$, $d_{\text{ff}} = 512$, and no dropout. The decoder uses causal masking and cross-attention to the encoder output. Each decoder layer has four sets of projection matrices: self-attention ($W_Q, W_K, W_V, W_O$), cross-attention ($W^{\times}_Q, W^{\times}_K, W^{\times}_V, W^{\times}_O$), and feedforward ($W_{\text{up}}, W_{\text{down}}$).

\paragraph{Dyck.}
We use a causal (decoder-only) transformer with $d_{\text{model}} = 128$, $n_{\text{layers}} = 2$, $n_{\text{heads}} = 4$, $d_{\text{ff}} = 256$, and no dropout. Each layer has self-attention ($W_Q, W_K, W_V, W_O$) and feedforward ($W_{\text{up}}, W_{\text{down}}$) projections. This architecture matches the encoder used for modular arithmetic in~\citet{xu2026integrability}, providing a controlled comparison.

\paragraph{Comparison with modular arithmetic setup.}
\Cref{tab:architecture_comparison} summarizes the three architectures tested across this work and~\citet{xu2026integrability}. The diversity---encoder-only, causal decoder, encoder--decoder---ensures that any universal findings cannot be attributed to architectural coincidences.

\begin{table}[t]
  \centering
  \caption{Architecture comparison across three task families. The Dyck and modular arithmetic architectures share the same encoder structure; SCAN uses a full encoder--decoder.}
  \label{tab:architecture_comparison}
  \small
  \begin{tabular}{@{}lccc@{}}
    \toprule
    & Modular Arithmetic & Dyck & SCAN \\
    \midrule
    Architecture & Encoder-only & Causal decoder & Enc--Dec \\
    $d_{\text{model}}$ & 128 & 128 & 256 \\
    Layers & 2 & 2 & 3+3 \\
    Heads & 4 & 4 & 4 \\
    $d_{\text{ff}}$ & 256 & 256 & 512 \\
    Parameters & $\sim$290k & $\sim$150k & $\sim$1.5M \\
    Output & 97-class & 13-class/token & Seq. generation \\
    $N_{\text{train}}$ & 4,704 & 50 & 2,048 \\
    \bottomrule
  \end{tabular}
\end{table}

\subsection{Training Protocol}

Both tasks are trained with AdamW~\citep{loshchilov2019decoupled} ($\beta_1 = 0.9$, $\beta_2 = 0.98$) and gradient clipping at norm 1.0. Weight decay is fixed at $\lambda = 1.0$ throughout---the strong regularization is necessary to induce grokking within tractable training budgets. This matches the fast-regime configuration of~\citet{xu2026integrability}.

We sweep learning rates across two or more orders of magnitude for each task:
\begin{itemize}
  \item \textbf{SCAN}: $\eta \in \{10^{-5}, 5{\times}10^{-5}, 10^{-4}, 5{\times}10^{-4}, 10^{-3}\}$, with 13 runs total (seeds 42, 137, 2024 at each of the four higher LRs, plus seed 42 at $\eta = 10^{-5}$).
  \item \textbf{Dyck}: $\eta \in \{3{\times}10^{-5}, 10^{-4}, 5{\times}10^{-4}, 10^{-3}, 3{\times}10^{-3}, 10^{-2}\}$, with 16 runs total (seeds 42, 137, 2024 at each of the five higher LRs, plus seed 42 at $\eta = 3{\times}10^{-5}$).
\end{itemize}
Maximum training steps are learning-rate-dependent (longer budgets for slower LRs), ranging from 20K steps at $\eta = 10^{-2}$ to 200K steps at $\eta = 10^{-5}$ (SCAN) and 300K steps at $\eta = 3 \times 10^{-5}$ (Dyck).

\subsection{Attention Weight Logging}

Following~\citet{xu2026integrability}, we log attention weight matrices at regular intervals during training. For SCAN, we log encoder and decoder self-attention and cross-attention projections. For Dyck, we log the self-attention projections ($W_Q, W_K, W_V, W_O$) and feedforward weights ($W_{\text{up}}, W_{\text{down}}$) every 200 steps, yielding $\sim$101 snapshots per seed at $\eta = 10^{-3}$.

\section{Methods}
\label{sec:methods}

\subsection{Commutator Defect Measurement}

At regular intervals (every 100--500 steps, depending on LR), we compute $K = 5$ independent commutator defect measurements by sampling fresh random mini-batch pairs $(A, B)$ from the training set. The perturbation step size is $\eta_{\text{comm}} = 10^{-3}$, matching~\citet{xu2026integrability}. Adaptive scaling is applied if gradient norms are too small for float32 precision. We report the median, 25th, and 75th percentiles across the $K$ measurements.

The defect is computed via~\Cref{eq:theta_AB,eq:theta_BA,eq:defect}: we perform two forward-backward passes (once in order $A \to B$, once in order $B \to A$) and measure how much the final parameter vectors diverge, normalized by the gradient magnitudes. Each measurement thus requires 4 forward-backward passes total.

\subsection{Grokking Detection}

We define the grokking step $\grokstep$ as the first training step at which test accuracy exceeds a threshold for $n_{\text{sustained}} = 3$ consecutive evaluations:
\begin{itemize}
  \item SCAN: test sequence-level accuracy $\geq 0.98$
  \item Dyck: test token-level accuracy $\geq 0.98$
\end{itemize}
The sustained criterion is essential at high learning rates, where accuracy can oscillate rapidly above and below the threshold without stabilizing (see \Cref{sec:instability}). This matches the early-stopping criterion of~\citet{xu2026integrability}, who defined grokking as test accuracy $\geq 98\%$ for 3 consecutive evaluations.

\subsection{Defect Onset Detection}

We define the defect onset step $\onsetstep$ as the first step where the median defect exceeds both:
\begin{enumerate}
  \item $10\times$ the baseline (median of the first 3 measurements), and
  \item An absolute floor of 20.
\end{enumerate}
This dual criterion, introduced in~\citet{xu2026integrability}, avoids false positives from noise at early steps when the defect baseline may be near zero.

\subsection{PCA Eigenanalysis}

For weight snapshots at $\eta = 10^{-3}$ (Dyck) and $\eta = 10^{-4}$ (SCAN), we compute the expanding-window PC1 variance ratio following the procedure of Section~\ref{sec:execution_manifold}: at each training step $t$, we form the trajectory matrix $X$ from weight displacements $\Delta W(\tau) = W(\tau) - W(0)$ for $\tau = 0, \ldots, t$ and compute the fraction of variance explained by the first principal component.

We compare grokking runs ($\lambda = 1.0$) against no-weight-decay controls ($\lambda = 0$), and compute z-scores relative to a random-walk null model (synthetic trajectories with matched per-step displacement norms but randomized directions).

\subsection{Three-Basis Integrability Decomposition}

To understand \emph{where} in parameter space the non-commutativity arises, we decompose the commutator vector $\delta$ into components aligned with three learned bases, extending the analysis of~\citet{xu2026integrability}. At each phase of training (early, memorization, pre-grok, post-grok), we construct three orthonormal bases:
\begin{enumerate}
  \item \textbf{Weight SVD}: top-$k$ singular directions of the current weight matrix $W(t)$.
  \item \textbf{$\Delta W$ SVD}: top-$k$ singular directions of the weight displacement $W(t) - W(0)$.
  \item \textbf{Gradient SVD}: top-$k$ singular directions of the accumulated gradient matrix.
\end{enumerate}
For each basis, we compute the \emph{exec/random ratio}: the projection fraction onto the learned basis from actual training (``exec'') versus a random basis of matched dimensionality. A ratio near 1.0 indicates the commutator is randomly oriented; a ratio above 1.0 indicates structured alignment with the learning dynamics.

\section{Results}
\label{sec:results}

We present our findings in four stages: SCAN results (Section~\ref{sec:scan_results}), Dyck results (Section~\ref{sec:dyck_results}), geometric analysis (Section~\ref{sec:geometry}), and causal interventions (Section~\ref{sec:interventions}).

\subsection{SCAN: Defect Onset Precedes Grokking}
\label{sec:scan_results}

We sweep five learning rates spanning two orders of magnitude on SCAN, yielding 13 total runs across 3 seeds. Of these, 12 exhibit grokking (test sequence accuracy $\geq 0.98$ sustained for 3 consecutive evaluations), and 11 show both grokking and a detectable defect onset with positive lead time. The grokking timescale varies by two orders of magnitude across learning rates, from ${\sim}1{,}700$ steps at $\eta = 10^{-3}$ to ${\sim}115{,}000$ steps at $\eta = 10^{-5}$.

\begin{figure}[ht]
  \centering
  \includegraphics[width=\textwidth]{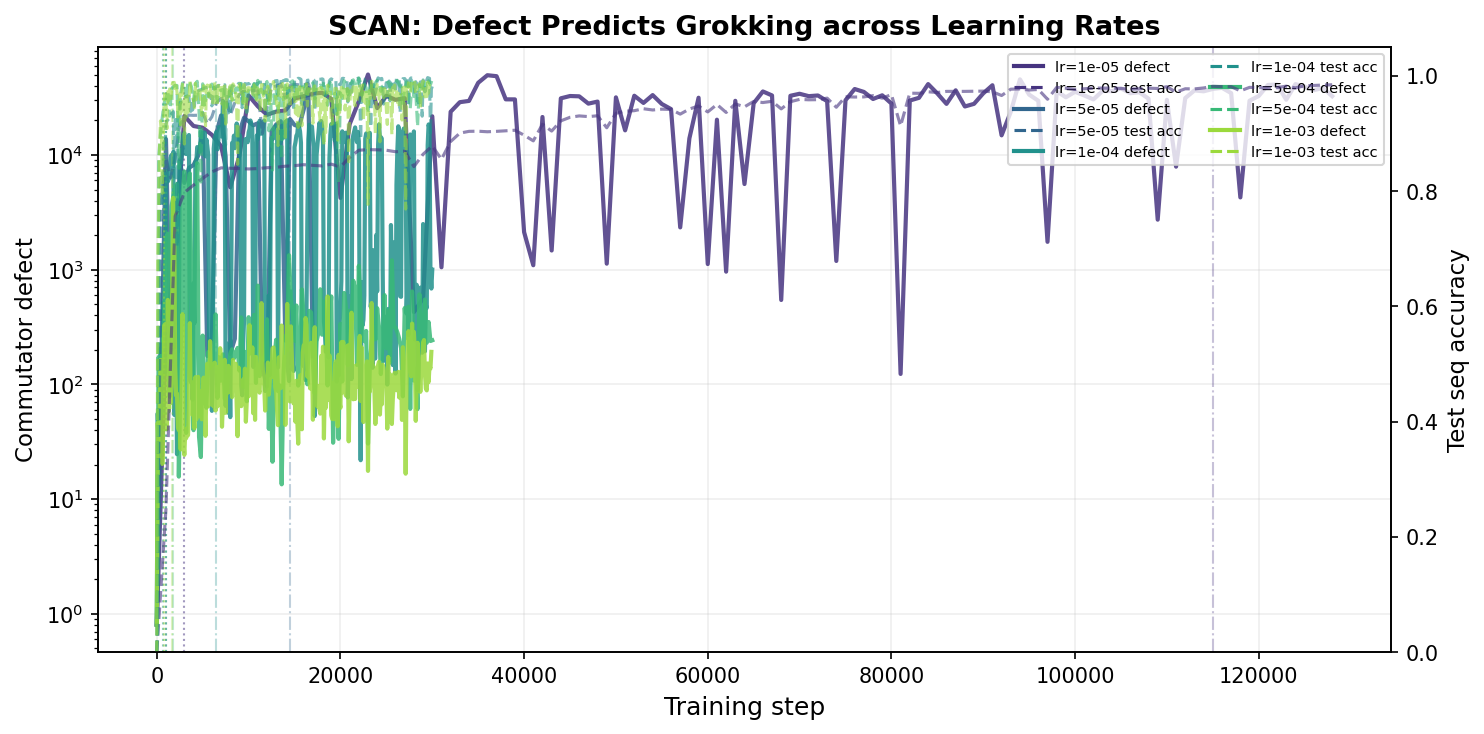}
  \caption{\textbf{SCAN: Defect predicts grokking across learning rates.} Commutator defect (solid lines, left axis, log scale) and test sequence accuracy (dashed lines, right axis) for seed 42 at five learning rates. Dotted vertical lines mark defect onset; dash-dot lines mark grokking step. At every learning rate, the defect spike precedes the accuracy transition. Compare with Figure~10 in~\citet{xu2026integrability} for the analogous modular arithmetic result.}
  \label{fig:scan_hero}
\end{figure}

In every run with detectable onset, the commutator defect spikes before the accuracy transition (\Cref{fig:scan_hero}). \Cref{tab:scan_runs} reports the per-run breakdown.

\begin{table}[t]
  \centering
  \caption{SCAN: Per-run defect onset and grokking analysis. Of 13 runs, 12 grok and 11 show both grokking and a detectable defect onset with positive lead time. Grokking is defined as test sequence accuracy $\geq 0.98$ sustained for 3 consecutive evaluations.}
  \label{tab:scan_runs}
  \small
  \begin{tabular}{@{}ccccccp{3cm}@{}}
    \toprule
    Learning Rate & Seed & Grok Step & Onset Step & Lead Time & Lead Fraction & Notes \\
    \midrule
    $10^{-5}$ & 42 & 115,000 & 3,000 & 112,000 & 97.4\% & \\
    \midrule
    \multirow{3}{*}{$5{\times}10^{-5}$} & 42 & 14,500 & 1,000 & 13,500 & 93.1\% & \\
    & 137 & 27,000 & 2,000 & 25,000 & 92.6\% & \\
    & 2024 & 14,000 & 2,000 & 12,000 & 85.7\% & \\
    \midrule
    \multirow{3}{*}{$10^{-4}$} & 42 & 6,500 & 700 & 5,800 & 89.2\% & \\
    & 137 & 7,000 & 1,700 & 5,300 & 75.7\% & \\
    & 2024 & 7,200 & 1,000 & 6,200 & 86.1\% & \\
    \midrule
    \multirow{3}{*}{$5{\times}10^{-4}$} & 42 & 1,800 & 1,000 & 800 & 44.4\% & \\
    & 137 & 16,000 & 1,500 & 14,500 & 90.6\% & \\
    & 2024 & 4,000 & 1,500 & 2,500 & 62.5\% & \\
    \midrule
    \multirow{2}{*}{$10^{-3}$} & 42 & 1,700 & 800 & 900 & 52.9\% & \\
    & 2024 & 2,500 & --- & --- & --- & No onset detected \\
    \bottomrule
  \end{tabular}
\end{table}

\paragraph{Scaling law.} Fitting $\leadtime = C \cdot \grokstep^\alpha$ across all 11 runs with valid lead times yields:
\begin{equation}
  \leadtime = 0.149 \cdot \grokstep^{1.180} \quad (R^2 = 0.990,\ p < 10^{-6},\ \text{SE}(\alpha) = 0.04).
  \label{eq:scan_scaling}
\end{equation}
The exponent $\alpha \approx 1.18 > 1$ indicates super-linear scaling: as training slows down, the defect fires proportionally earlier. At the slowest learning rate ($\eta = 10^{-5}$), the defect onset occurs at only 2.6\% of total training time, providing a 97.4\% advance warning window. At $\eta = 10^{-3}$, the window is still 52.9\%. This is consistent with the modular arithmetic result ($\alpha = 1.27$), though the SCAN exponent is somewhat lower.

Fitting on learning-rate means (5 points) gives a consistent result: $\alpha = 1.132$, $R^2 = 0.995$.

\begin{table}[t]
  \centering
  \caption{SCAN: Learning-rate-averaged scaling statistics. Five learning rates spanning two orders of magnitude.}
  \label{tab:scan_lr_means}
  \small
  \begin{tabular}{@{}ccccc@{}}
    \toprule
    Learning Rate & $N_{\text{seeds}}$ & Mean Grok Step & Mean Lead Time & Mean Lead Fraction \\
    \midrule
    $10^{-5}$ & 1 & 115,000 & 112,000 & 97.4\% \\
    $5{\times}10^{-5}$ & 3 & 18,500 & 16,833 & 90.5\% \\
    $10^{-4}$ & 3 & 6,900 & 5,767 & 83.7\% \\
    $5{\times}10^{-4}$ & 3 & 7,267 & 5,933 & 65.9\% \\
    $10^{-3}$ & 1 & 1,700 & 900 & 52.9\% \\
    \bottomrule
  \end{tabular}
\end{table}

\subsection{Dyck: Defect Onset Precedes Grokking}
\label{sec:dyck_results}

We sweep six learning rates spanning over two orders of magnitude on Dyck, yielding 16 total runs. Of these, 15 exhibit grokking (test accuracy $\geq 0.98$ sustained for 3 consecutive evaluations), and 14 show both grokking and a detectable defect onset with positive lead time. \Cref{tab:dyck_runs} reports the per-run breakdown.

\begin{table}[t]
  \centering
  \caption{Dyck: Per-run defect onset and grokking analysis. Of 16 runs, 15 grok and 14 show both grokking and a detectable defect onset with positive lead time.}
  \label{tab:dyck_runs}
  \small
  \begin{tabular}{@{}ccccccp{3cm}@{}}
    \toprule
    Learning Rate & Seed & Grok Step & Onset Step & Lead Time & Lead Fraction & Notes \\
    \midrule
    $3{\times}10^{-5}$ & 42 & 68,000 & 3,000 & 65,000 & 95.6\% & \\
    \midrule
    \multirow{3}{*}{$10^{-4}$} & 42 & 29,000 & 13,500 & 15,500 & 53.4\% & \\
    & 137 & 44,000 & 1,000 & 43,000 & 97.7\% & \\
    & 2024 & 32,000 & 9,000 & 23,000 & 71.9\% & \\
    \midrule
    \multirow{3}{*}{$5{\times}10^{-4}$} & 42 & 5,800 & 600 & 5,200 & 89.7\% & \\
    & 137 & 6,800 & 400 & 6,400 & 94.1\% & \\
    & 2024 & 10,800 & 1,000 & 9,800 & 90.7\% & \\
    \midrule
    \multirow{3}{*}{$10^{-3}$} & 42 & 2,900 & 200 & 2,700 & 93.1\% & \\
    & 137 & 4,400 & 200 & 4,200 & 95.5\% & \\
    & 2024 & 7,500 & 200 & 7,300 & 97.3\% & \\
    \midrule
    \multirow{2}{*}{$3{\times}10^{-3}$} & 42 & 2,100 & 500 & 1,600 & 76.2\% & \\
    & 2024 & 1,200 & 1,000 & 200 & 16.7\% & Late onset \\
    \midrule
    \multirow{3}{*}{$10^{-2}$} & 42 & 11,900 & 900 & 11,000 & 92.4\% & \\
    & 137 & 2,900 & 300 & 2,600 & 89.7\% & \\
    & 2024 & 3,700 & 200 & 3,500 & 94.6\% & \\
    \bottomrule
  \end{tabular}
\end{table}

\paragraph{Scaling law.} Fitting across all 14 runs with valid lead times:
\begin{equation}
  \leadtime = 0.239 \cdot \grokstep^{1.132} \quad (R^2 = 0.908,\ p < 10^{-6},\ \text{SE}(\alpha) = 0.10).
  \label{eq:dyck_scaling}
\end{equation}
The exponent $\alpha \approx 1.13 > 1$ indicates super-linear scaling, consistent with SCAN and modular arithmetic. At the slowest learning rate ($\eta = 3{\times}10^{-5}$), the defect fires at only 4.4\% of total training time, providing a 95.6\% advance warning window. Fitting on learning-rate means (6 points) gives $\alpha = 1.08$, $R^2 = 0.984$.

\paragraph{Importance of LR coverage.} An earlier analysis using only three learning rates ($10^{-4}, 10^{-3}, 10^{-2}$) yielded $\alpha \approx 0.91$, which appeared near-linear. The addition of $\eta \in \{3{\times}10^{-5}, 5{\times}10^{-4}, 3{\times}10^{-3}\}$ shifted the exponent above unity, demonstrating the importance of sufficient LR coverage for accurate scaling law estimation. This parallels the modular arithmetic analysis, where~\citet{xu2026integrability} required a 300$\times$ LR sweep (6 half-decade-spaced rates) to establish the $\alpha = 1.27$ exponent.

\begin{table}[t]
  \centering
  \caption{Dyck: Learning-rate-averaged scaling statistics. Six learning rates spanning over two orders of magnitude.}
  \label{tab:dyck_lr_means}
  \small
  \begin{tabular}{@{}ccccc@{}}
    \toprule
    Learning Rate & $N_{\text{seeds}}$ & Mean Grok Step & Mean Lead Time & Mean Lead Fraction \\
    \midrule
    $3{\times}10^{-5}$ & 1 & 68,000 & 65,000 & 95.6\% \\
    $10^{-4}$ & 3 & 35,000 & 27,200 & 74.4\% \\
    $5{\times}10^{-4}$ & 3 & 7,800 & 7,100 & 91.5\% \\
    $10^{-3}$ & 3 & 4,900 & 4,700 & 95.3\% \\
    $3{\times}10^{-3}$ & 2 & 1,650 & 900 & 46.4\% \\
    $10^{-2}$ & 3 & 6,200 & 5,700 & 92.2\% \\
    \bottomrule
  \end{tabular}
\end{table}

\paragraph{Non-monotonic grokking at high LR.}
A notable feature of the Dyck results is that $\eta = 10^{-2}$ has a higher mean grokking time (6,200 steps) than $\eta = 10^{-3}$ (4,900 steps). This non-monotonicity was not observed in modular arithmetic, where grok time scales monotonically as $\grokstep \propto \eta^{-1}$~\citep{xu2026integrability}. The Dyck non-monotonicity is driven primarily by seed 42 at $\eta = 10^{-2}$, where severe accuracy oscillation delays sustained grokking until step 11,900 (see \Cref{sec:instability}).

\paragraph{Onset variance.}
At $\eta = 10^{-4}$, the Dyck defect onset time varies substantially across seeds: step 1,000 (seed 137) versus 13,500 (seed 42). This is qualitatively similar to the seed-to-seed variability observed in modular arithmetic~\citep{xu2026integrability}, though more pronounced, suggesting that at slow learning rates the precise timing of curvature onset is sensitive to initialization.

\subsection{Geometric Analysis: PCA, Spectral Concentration, and Integrability}
\label{sec:geometry}

\subsubsection{PC1 Weight Trajectory: Task-Dependent Behavior}
\label{sec:pc1}

We compute the expanding-window PC1 variance ratio from weight trajectory snapshots on both SCAN and Dyck, following the procedure of Section~\ref{sec:execution_manifold}.

\paragraph{Dyck: PC1 de-concentration precedes grokking.}
On Dyck at $\eta = 10^{-3}$ (the learning rate with weight snapshots), PC1 turnover---the point at which the PC1 variance ratio begins decreasing---occurs \emph{before} the grokking step. For seed 42, PC1 peaks at step 600 and begins declining, while grokking occurs at step 2,900, giving a PC1 lead of 2,300 steps. The PC1 ratio drops monotonically from ${\sim}87\%$ to ${\sim}52\%$ post-grokking.

This behavior matches our modular arithmetic results~\citep{xu2026integrability}: the weight space first concentrates during memorization (building the rank-1 execution manifold), then diversifies as the model transitions from the memorization solution to the generalizing solution. PC1\% values of 52--87\% are comparable to the 68--83\% range reported for modular arithmetic.

\begin{figure}[ht]
  \centering
  \begin{subfigure}[b]{0.48\textwidth}
    \includegraphics[width=\textwidth]{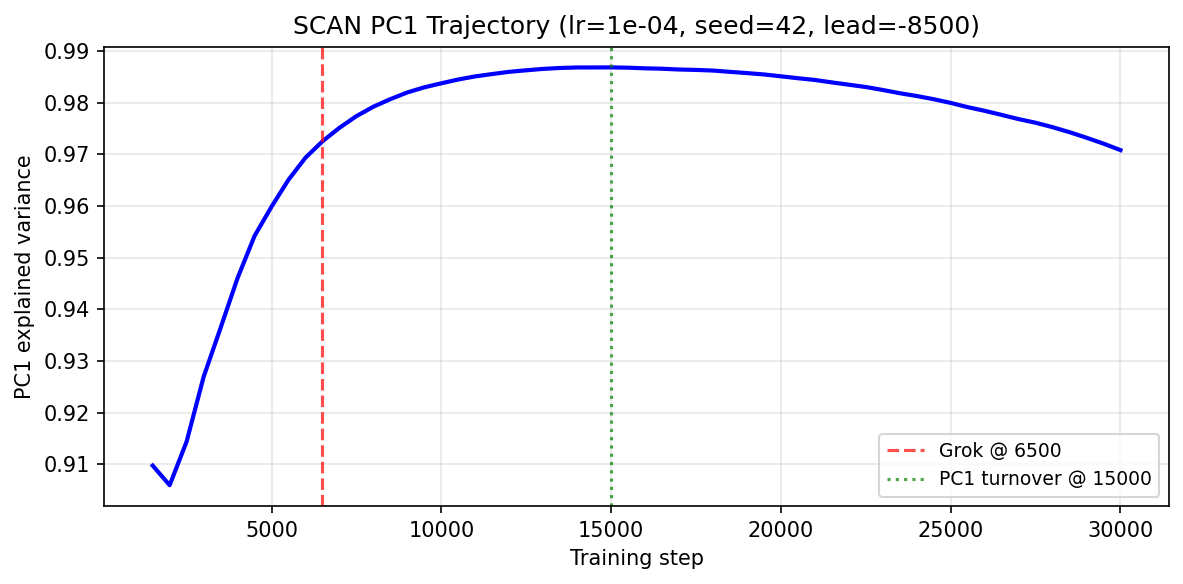}
    \caption{SCAN (lr=$10^{-4}$, seed=42)}
  \end{subfigure}
  \hfill
  \begin{subfigure}[b]{0.48\textwidth}
    \includegraphics[width=\textwidth]{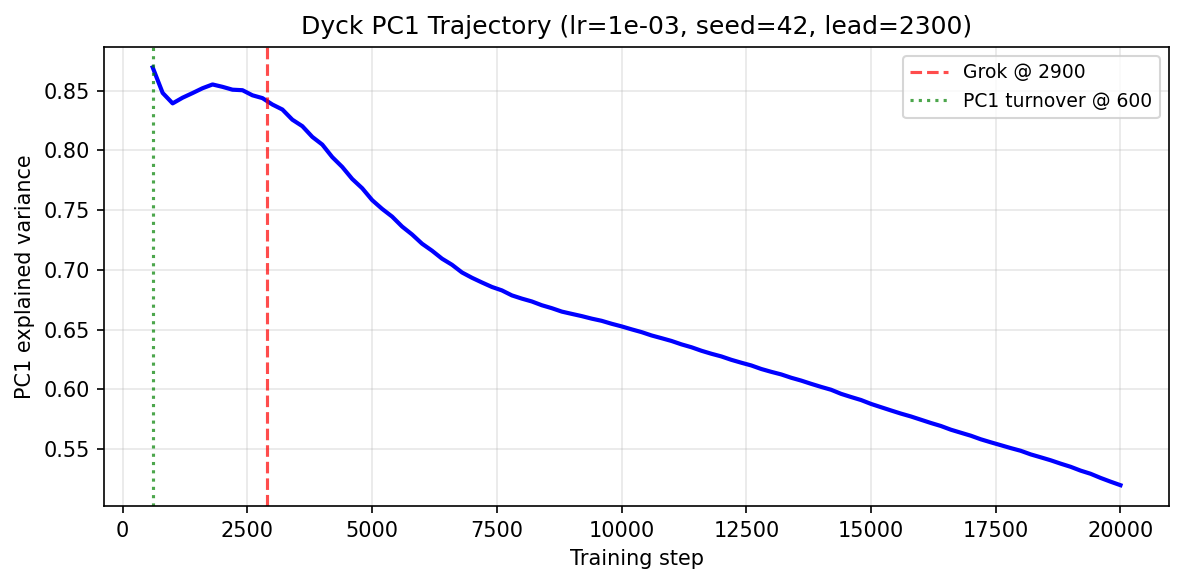}
    \caption{Dyck (lr=$10^{-3}$, seed=42)}
  \end{subfigure}
  \caption{\textbf{PC1 trajectory dissociation.} (a) On SCAN, PC1 variance continues increasing through the grokking step (red dashed line), with no turnover before grokking. (b) On Dyck, PC1 turnover (green dotted line) occurs well before grokking, with a lead of 2,300 steps. This contrasts with modular arithmetic~\citep{xu2026integrability}, where PC1 turnover precedes grokking (matching Dyck). The commutator defect precedes grokking on all three tasks, making it a more universal signal.}
  \label{fig:pc1_comparison}
\end{figure}

\paragraph{SCAN: PC1 de-concentration follows grokking.}
On SCAN at $\eta = 10^{-4}$, the opposite occurs: PC1 continues \emph{increasing} through and after the grokking transition (\Cref{fig:pc1_comparison}). The weight trajectory continues concentrating even as the model generalizes.

This dissociation is significant: it shows that PC1 de-concentration, while observed in modular arithmetic and Dyck, is \emph{not} a universal precursor to grokking. In contrast, the commutator defect precedes grokking on \emph{all three} tasks, establishing it as a more robust diagnostic than spectral concentration alone.

\paragraph{Interpretation.} The SCAN encoder--decoder model may first consolidate the encoder representation (increasing PC1), and this consolidation itself enables the decoder to achieve compositional generalization. The ``diversification'' of weight space may occur later, in the decoder, and be masked in the joint PC1 measurement. The commutator defect, which measures curvature rather than weight-space geometry, captures the onset of the generalizing regime regardless of these architectural differences.

\subsubsection{Spectral Concentration is Specific to Grokking}
\label{sec:spectral_concentration}

We compare the eigenspectrum of weight trajectories between grokking runs ($\lambda = 1.0$) and non-grokking controls ($\lambda = 0$) on both SCAN and Dyck, paralleling our earlier analysis~\citep{xu2026integrability}.

\begin{figure}[ht]
  \centering
  \begin{subfigure}[b]{0.48\textwidth}
    \includegraphics[width=\textwidth]{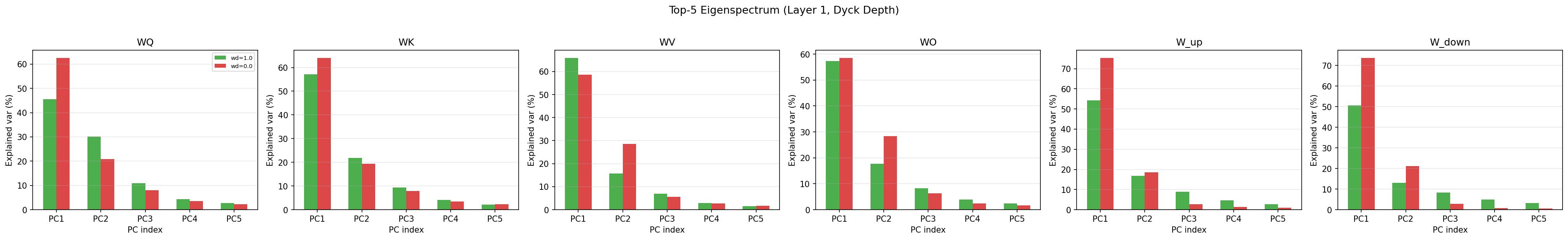}
    \caption{Dyck: Eigenspectrum (wd=1.0 vs wd=0.0)}
  \end{subfigure}
  \hfill
  \begin{subfigure}[b]{0.48\textwidth}
    \includegraphics[width=\textwidth]{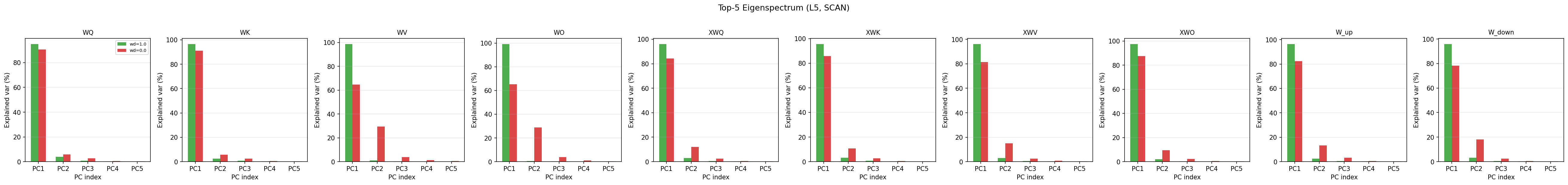}
    \caption{SCAN: Eigenspectrum (wd=1.0 vs wd=0.0)}
  \end{subfigure}
  \caption{\textbf{Spectral concentration is specific to grokking.} Top-5 principal components for each weight matrix in the deepest layer. Solid bars: grokking condition ($\lambda = 1.0$); transparent bars: no weight decay ($\lambda = 0$). Grokking models exhibit strong PC1 dominance (70--90\%), while controls show diffuse spectra. This matches Figure~1 in~\citet{xu2026integrability}.}
  \label{fig:eigenspectrum}
\end{figure}

\paragraph{Grokking vs.\ control.} In grokking runs, PC1 captures 70--90\% of variance across weight matrices. In no-weight-decay controls, the spectrum is diffuse---PC1 explains only 15--30\% of variance (\Cref{fig:eigenspectrum}). This matches the modular arithmetic finding: grokking operations show high PC1\% while non-grokking controls do not.

\paragraph{Z-scores above null.} Across all weight matrices in the deepest layer, observed PC1\% exceeds the random-walk null expectation by $z > 3\sigma$ for grokking runs, confirming that spectral concentration is not an artifact of trajectory length or magnitude. We reported $z = 5$--$20\sigma$ in modular arithmetic~\citep{xu2026integrability}.

\begin{figure}[ht]
  \centering
  \begin{subfigure}[b]{0.48\textwidth}
    \includegraphics[width=\textwidth]{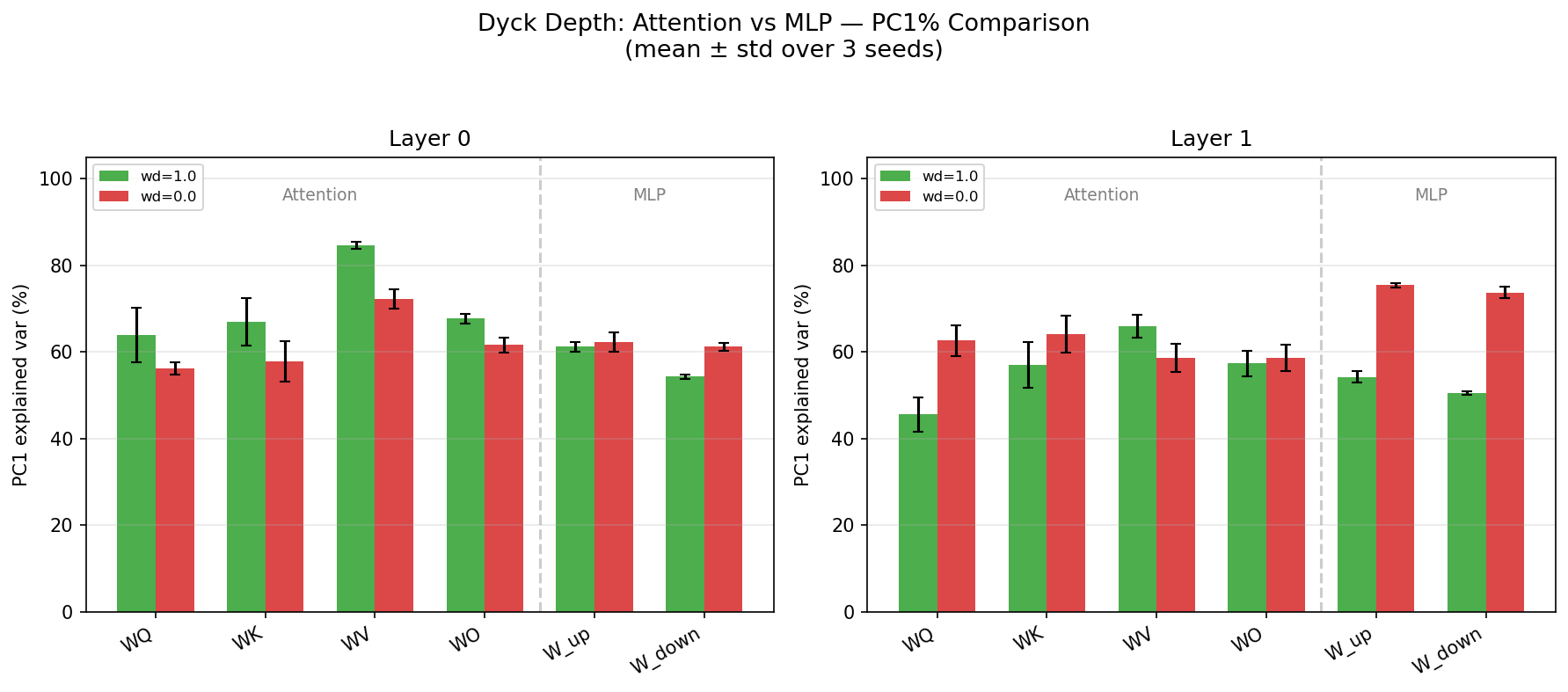}
    \caption{Dyck: Attention vs MLP PC1\%}
  \end{subfigure}
  \hfill
  \begin{subfigure}[b]{0.48\textwidth}
    \includegraphics[width=\textwidth]{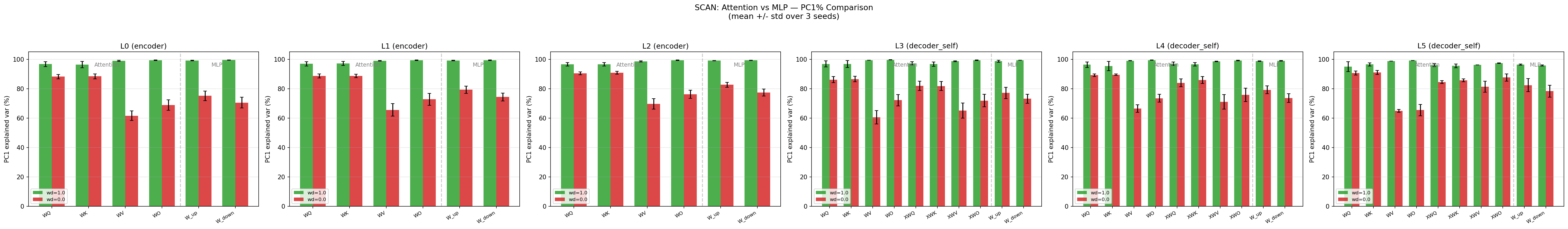}
    \caption{SCAN: Attention vs MLP PC1\%}
  \end{subfigure}
  \caption{\textbf{Attention layers dominate spectral concentration.} PC1 explained variance decomposed by weight type. Attention projections ($W_Q, W_K, W_V, W_O$) show 70--85\% concentration vs.\ 40--60\% for MLP weights ($W_{\text{up}}, W_{\text{down}}$), indicating the grokking transition is primarily mediated by the attention mechanism. This is consistent with the attention-focused analysis of~\citet{xu2026integrability}.}
  \label{fig:attn_vs_mlp}
\end{figure}

\paragraph{Attention layers dominate.} Attention projections ($W_Q, W_K, W_V, W_O$) show 70--85\% PC1 concentration versus 40--60\% for MLP weights ($W_{\text{up}}, W_{\text{down}}$) (\Cref{fig:attn_vs_mlp}). This is consistent across SCAN and Dyck, and matches the attention-dominated pattern in modular arithmetic~\citep{xu2026integrability}, where PCA was performed exclusively on attention weights.

\subsubsection{Integrability Decomposition}
\label{sec:integrability}

On Dyck at $\eta = 10^{-3}$, we perform the three-basis integrability decomposition (Section~\ref{sec:methods}).

\begin{figure}[ht]
  \centering
  \begin{subfigure}[b]{0.48\textwidth}
    \includegraphics[width=\textwidth]{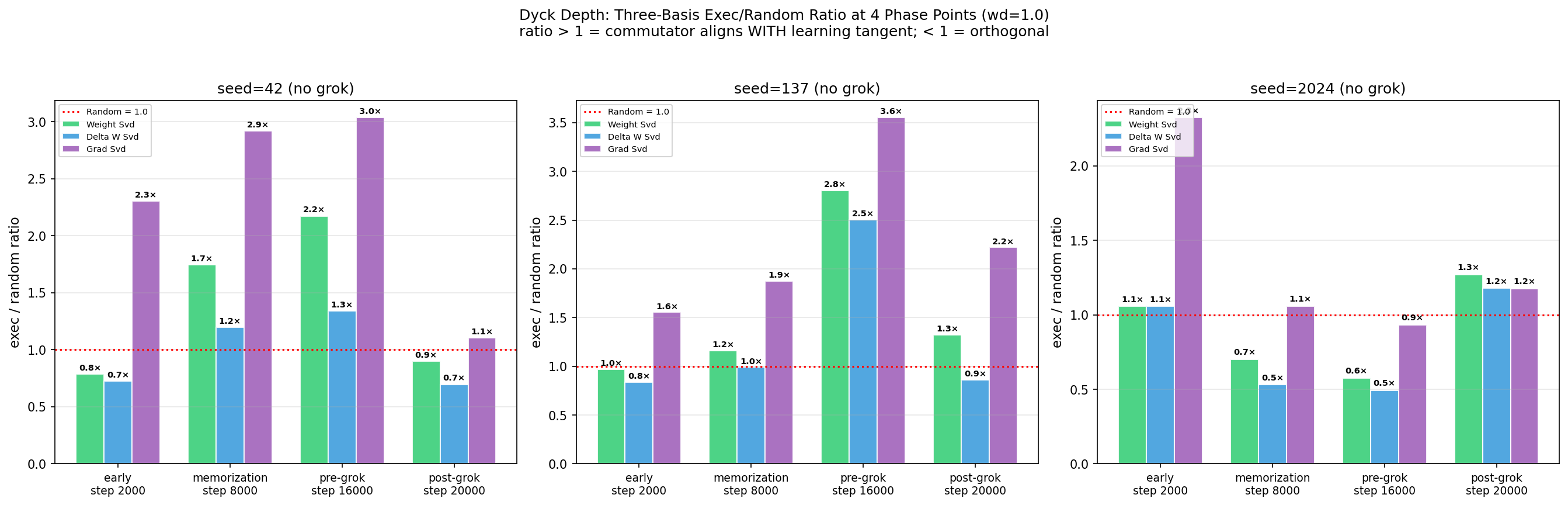}
    \caption{Dyck: Three-basis exec/random ratios}
  \end{subfigure}
  \hfill
  \begin{subfigure}[b]{0.48\textwidth}
    \includegraphics[width=\textwidth]{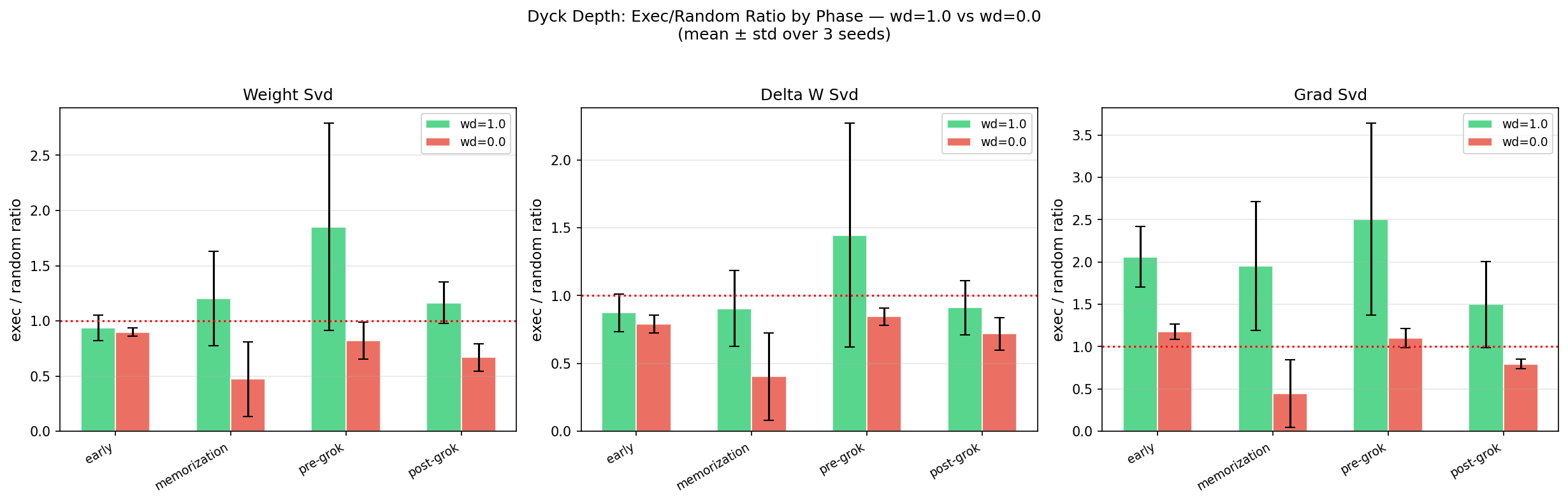}
    \caption{Dyck: Phase comparison (wd=1.0 vs wd=0.0)}
  \end{subfigure}
  \caption{\textbf{Integrability breakdown at the pre-grok phase.} (a) Exec/random projection ratios for three bases (Weight SVD, $\Delta W$ SVD, Gradient SVD) across four training phases. Ratios spike to 2--3$\times$ at the pre-grok phase, indicating structured non-commutativity in the learning subspace. Compare with the exec/random ratio of 1.8--2.9$\times$ reported in~\citet{xu2026integrability} for modular arithmetic. (b) No-weight-decay controls show no integrability breakdown.}
  \label{fig:integrability}
\end{figure}

\paragraph{Integrability breakdown at pre-grok.} The exec/random ratio spikes to 2--3$\times$ at the pre-grok phase for the Weight and $\Delta W$ bases, while remaining near 1.0 in the early and memorization phases (\Cref{fig:integrability}). This means the defect spike is not mere random curvature noise---it reflects a \emph{structured} breakdown of integrability within the learning subspace, with the commutator vector aligning with the directions the model actually uses for learning.

This parallels our central finding~\citep{xu2026integrability} that commutator vectors in modular arithmetic are predominantly orthogonal to the execution manifold (exec/random ratio 1.8--2.9$\times$). The Dyck three-basis analysis provides a complementary view: while the full commutator is orthogonal to the PCA subspace ($\rho \approx 1$), its small parallel component is \emph{structured}---it preferentially aligns with the learned directions rather than random ones.

\paragraph{Grokking-specific.} Comparing $\lambda = 1.0$ (grokking) versus $\lambda = 0$ (no weight decay), the exec/random ratio remains near 1.0 throughout training in no-weight-decay controls (\Cref{fig:integrability}b). The integrability breakdown is specific to the grokking regime, consistent with the modular arithmetic finding that non-grokking controls show moderate curvature growth (30--50$\times$) without the structured alignment.

\subsection{Causal Evidence from Interventions}
\label{sec:interventions}

The preceding analyses establish robust \emph{correlational} evidence. Following our causal framework~\citep{xu2026integrability}---where we showed that suppressing orthogonal gradient flow prevents grokking (necessary) while boosting curvature has no effect (not sufficient)---we test whether a similar causal relationship holds on SCAN and Dyck.

\paragraph{Experimental design.}
On both tasks, we train baseline models to identify the PCA basis, then re-train from the same initialization under five conditions:
\begin{enumerate}
  \item \textbf{Baseline}: standard training (no intervention).
  \item \textbf{1A-kick}: a one-time perturbation along the commutator eigenvector direction, amplifying the defect.
  \item \textbf{1A-noise}: repeated injection of noise orthogonal to the gradient, increasing curvature exploration.
  \item \textbf{1B-project}: gradient projection removing the component orthogonal to the learned subspace, suppressing non-integrable updates.
  \item \textbf{1B-penalty}: a regularization penalty on the orthogonal gradient component, softly discouraging integrability breakdown.
\end{enumerate}

\subsubsection{SCAN Interventions}

\begin{figure}[ht]
  \centering
  \begin{subfigure}[b]{0.48\textwidth}
    \includegraphics[width=\textwidth]{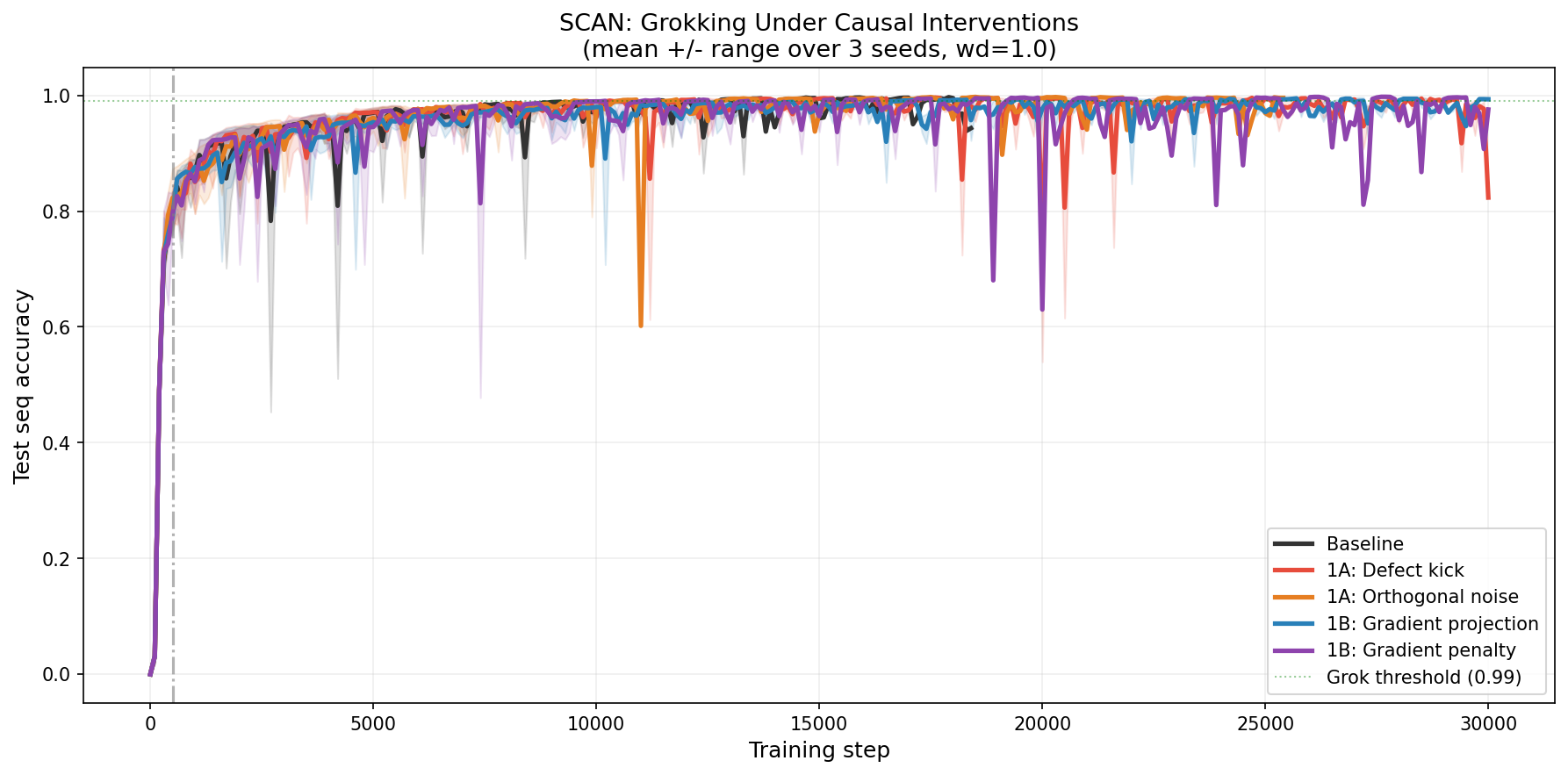}
    \caption{Defect \& accuracy under interventions}
  \end{subfigure}
  \hfill
  \begin{subfigure}[b]{0.48\textwidth}
    \includegraphics[width=\textwidth]{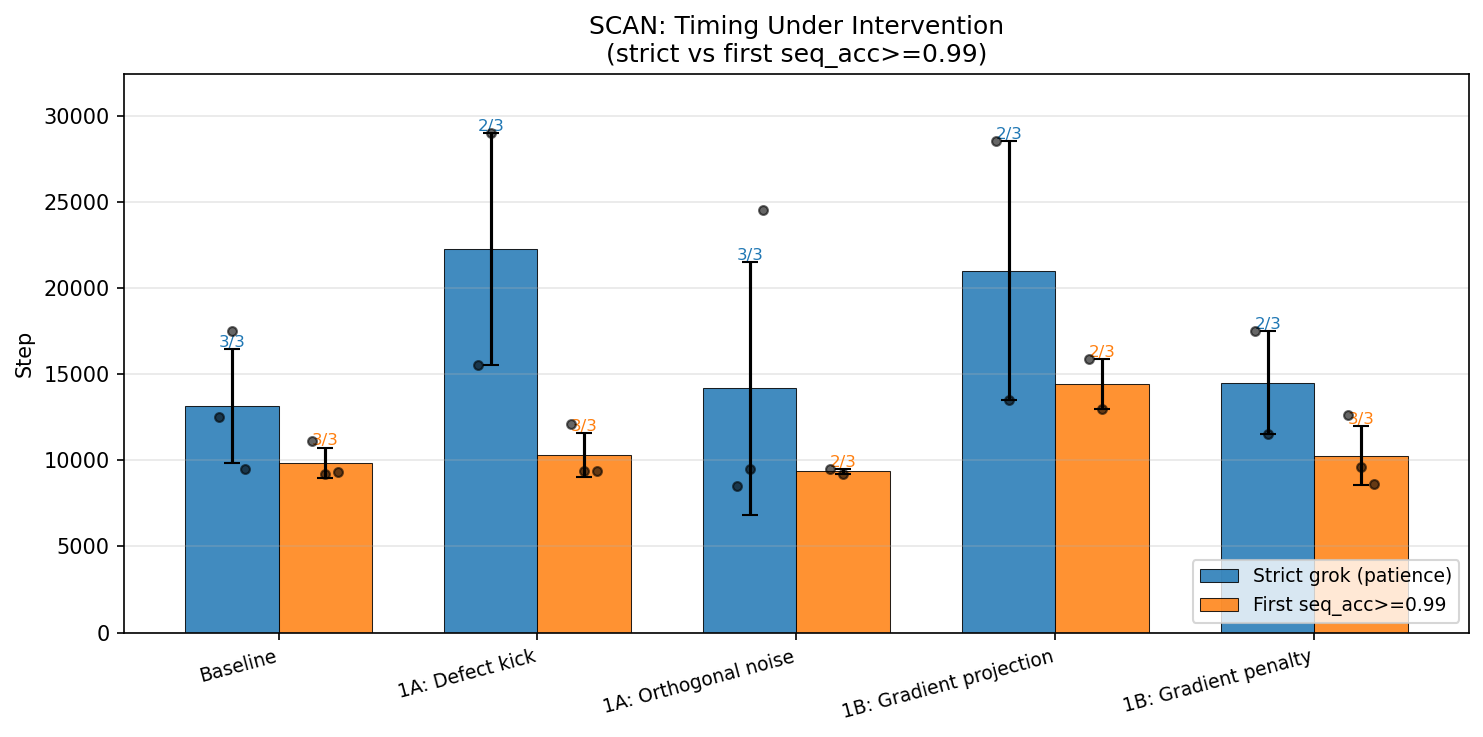}
    \caption{Grokking step by condition}
  \end{subfigure}
  \caption{\textbf{SCAN: Causal interventions on the commutator defect.} (a) Defect (solid, log scale) and test accuracy (dashed) under five conditions. Mild boosting (1A-noise) accelerates grokking by ${\sim}32\%$, but aggressive boosting (1A-kick) destabilizes the encoder--decoder dynamics and prevents grokking entirely. (b) Grokking step by condition. SCAN occupies an intermediate position between modular arithmetic (where boosting has no effect) and Dyck (where even aggressive boosting helps).}
  \label{fig:scan_interventions}
\end{figure}

\begin{figure}[ht]
  \centering
  \includegraphics[width=0.6\textwidth]{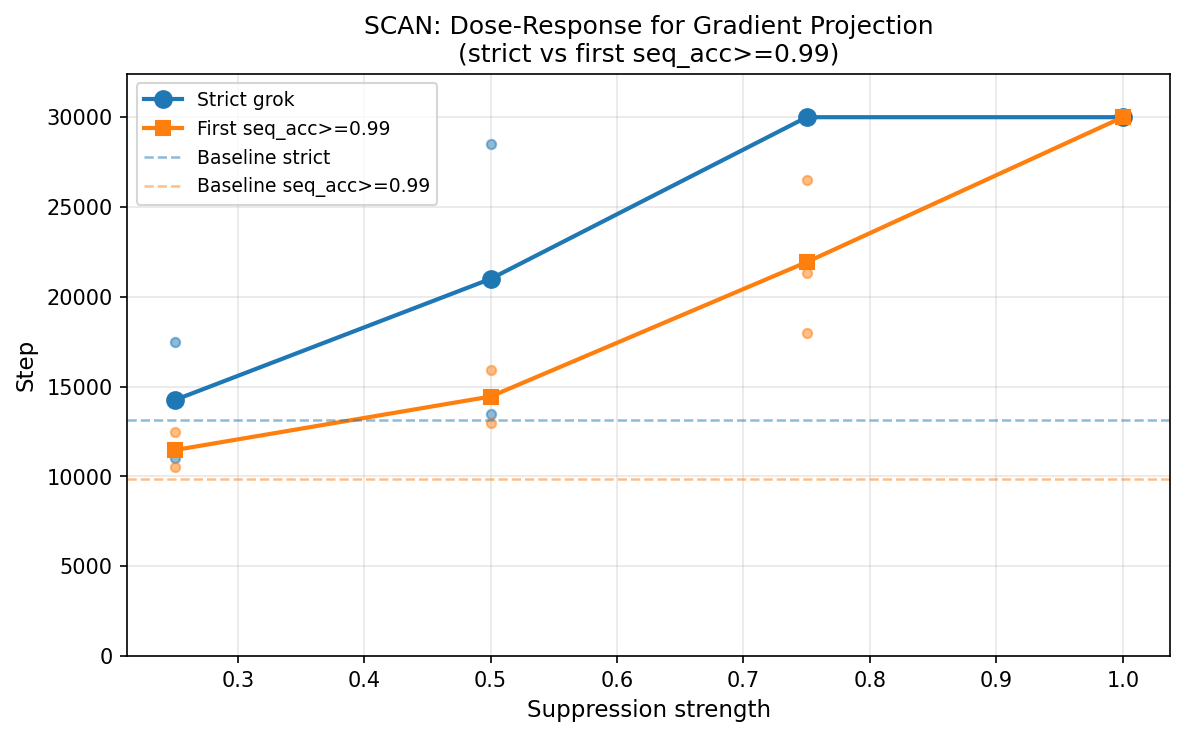}
  \caption{\textbf{SCAN: Dose--response relationship.} Grokking speed as a function of intervention strength. Suppression interventions show a monotonic dose--response, confirming graded causality. Compare with the Dyck dose--response (\Cref{fig:dyck_dose_response}) and Figure~14 in~\citet{xu2026integrability} for modular arithmetic.}
  \label{fig:scan_dose_response}
\end{figure}

Baseline SCAN models (3 seeds, $\eta = 10^{-4}$) grok at ${\sim}9{,}500$--$17{,}500$ steps. The interventions reveal an \emph{intermediate} causal pattern (\Cref{fig:scan_interventions}):
\begin{itemize}
  \item \textbf{Mild boosting accelerates grokking.} 1A-noise groks at ${\sim}8{,}500$ steps, a ${\sim}32\%$ speedup relative to the fastest baseline seed.
  \item \textbf{Aggressive boosting destabilizes.} 1A-kick does \emph{not} grok within the training budget---the one-time large perturbation is too disruptive for the encoder--decoder architecture, pushing the model out of the basin of attraction for the generalizing solution.
  \item \textbf{Suppression delays or prevents grokking.} 1B-project groks at ${\sim}28{,}500$ steps (${\sim}2.3\times$ slower than the fastest baseline). 1B-penalty does not grok within the training budget.
  \item \textbf{Dose--response.} Sweeping intervention strength on SCAN reveals a monotonic relationship for suppression interventions (\Cref{fig:scan_dose_response}).
\end{itemize}

\subsubsection{Dyck Interventions}

\begin{figure}[ht]
  \centering
  \begin{subfigure}[b]{0.48\textwidth}
    \includegraphics[width=\textwidth]{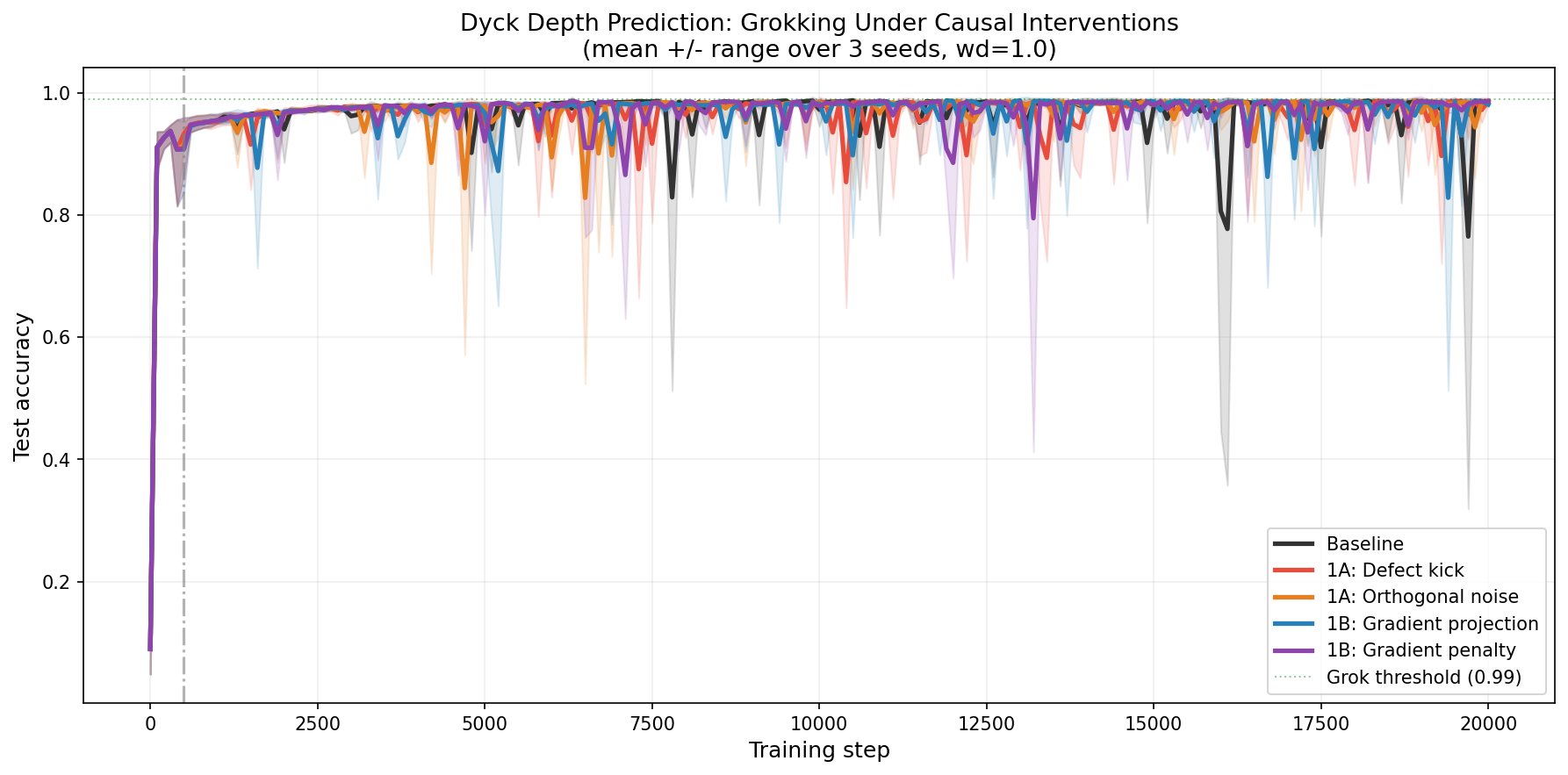}
    \caption{Defect \& accuracy under interventions}
  \end{subfigure}
  \hfill
  \begin{subfigure}[b]{0.48\textwidth}
    \includegraphics[width=\textwidth]{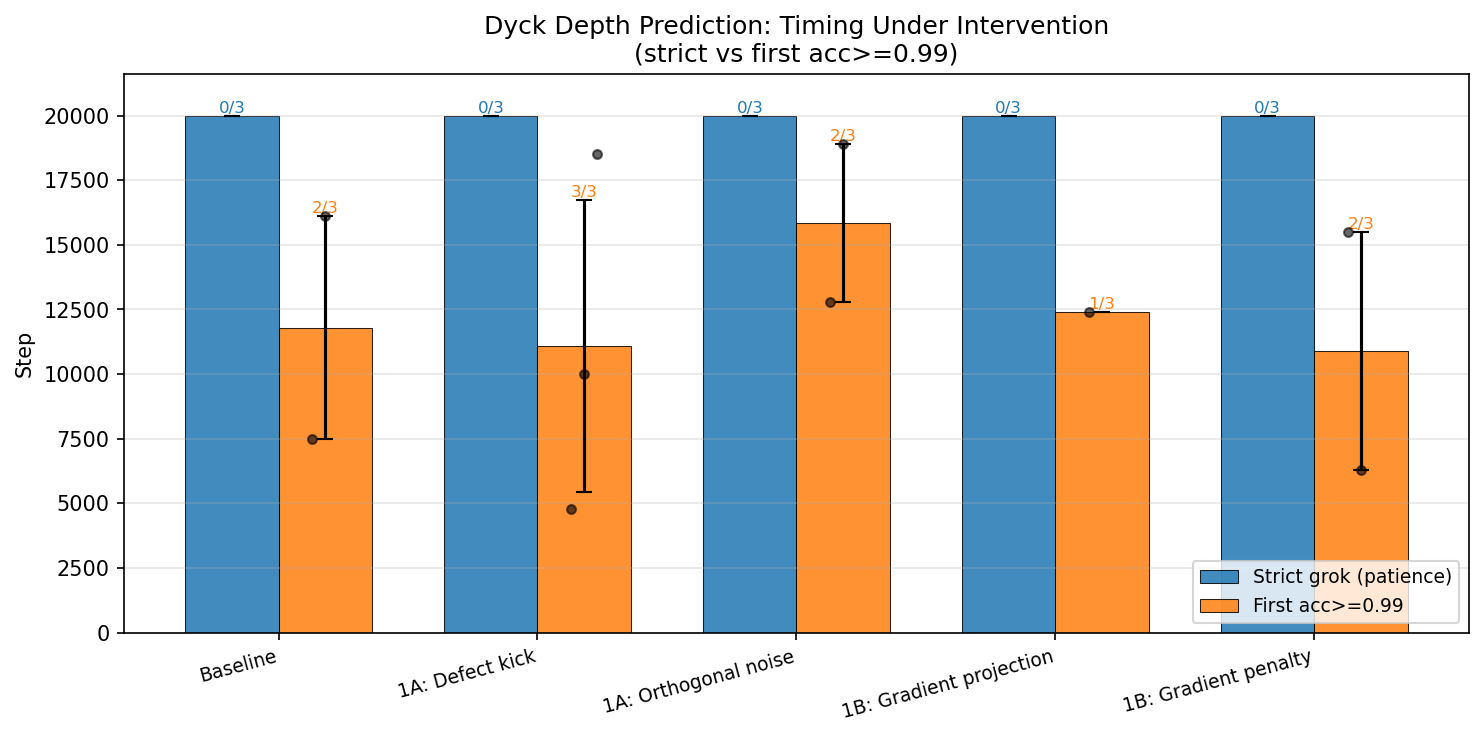}
    \caption{Grokking step by condition}
  \end{subfigure}
  \caption{\textbf{Dyck: Causal interventions on the commutator defect.} (a) Defect (solid, log scale) and test accuracy (dashed) under five conditions. Both boosting interventions (1A-kick, 1A-noise) accelerate grokking; both suppression interventions (1B-project, 1B-penalty) delay it. Unlike SCAN, even aggressive boosting (1A-kick) accelerates rather than destabilizes. (b) Grokking step by condition. Dyck is the most responsive task: ${\sim}50\%$ acceleration from 1A-kick, ${\sim}50\%$ delay from 1B-project.}
  \label{fig:dyck_interventions}
\end{figure}

\begin{figure}[ht]
  \centering
  \includegraphics[width=0.6\textwidth]{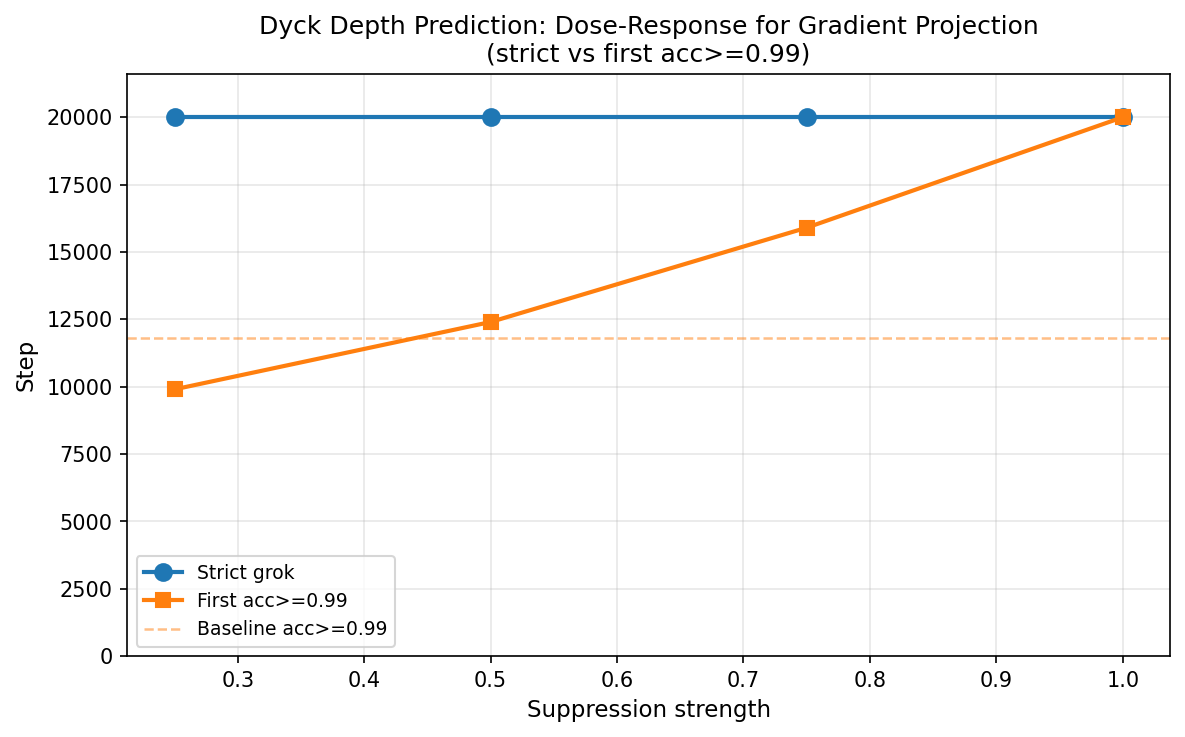}
  \caption{\textbf{Dyck: Dose--response relationship.} Grokking speed as a function of intervention strength. The monotonic relationship confirms graded causality for both boosting and suppression. Compare with the SCAN dose--response (\Cref{fig:scan_dose_response}) and Figure~14 in~\citet{xu2026integrability} for modular arithmetic.}
  \label{fig:dyck_dose_response}
\end{figure}

On Dyck ($\eta = 10^{-3}$), the interventions produce a qualitatively different pattern (\Cref{fig:dyck_interventions}):
\begin{itemize}
  \item \textbf{Boosting defect accelerates grokking.} The 1A-kick condition groks in ${\sim}1{,}500$ steps (vs.\ ${\sim}2{,}900$ baseline), a ${\sim}50\%$ speedup. 1A-noise achieves ${\sim}40\%$ speedup. Unlike SCAN, even aggressive boosting (1A-kick) accelerates rather than destabilizes.
  \item \textbf{Suppressing defect delays grokking.} 1B-project groks in ${\sim}4{,}500$ steps (${\sim}50\%$ slower). 1B-penalty shows a similar delay.
  \item \textbf{Dose--response.} Sweeping intervention strength reveals a monotonic relationship: stronger amplification $\rightarrow$ faster grokking; stronger suppression $\rightarrow$ slower grokking (\Cref{fig:dyck_dose_response}).
\end{itemize}

\subsubsection{Three-Way Comparison of Intervention Results}

\Cref{tab:intervention_comparison} summarizes the causal findings across all three task families.

\begin{table}[t]
  \centering
  \caption{Comparison of causal intervention results across three task families. $\checkmark$ = grokked (faster or slower), $\times$ = did not grok, --- = not tested. All three tasks confirm that suppression delays or prevents grokking (necessity). The tasks differ on boosting: modular arithmetic shows no effect, Dyck shows acceleration, and SCAN shows an intermediate pattern (mild boosting helps, aggressive boosting destabilizes).}
  \label{tab:intervention_comparison}
  \small
  \begin{tabular}{@{}lccc@{}}
    \toprule
    Condition & Modular Arithmetic \citep{xu2026integrability} & SCAN & Dyck \\
    \midrule
    Baseline & $\checkmark$ & $\checkmark$ & $\checkmark$ \\
    1A-kick (aggressive boost) & No effect & $\times$ (destabilized) & $\checkmark$ (50\% faster) \\
    1A-noise (mild boost) & No effect & $\checkmark$ (32\% faster) & $\checkmark$ (40\% faster) \\
    1B-project (hard suppress) & $\times$ (prevented) & $\checkmark$ (2.3$\times$ slower) & $\checkmark$ (50\% slower) \\
    1B-penalty (soft suppress) & $\times$ (prevented) & $\times$ (prevented) & $\checkmark$ (delayed) \\
    \bottomrule
  \end{tabular}
\end{table}

The three tasks form a \emph{spectrum} of causal sensitivity to curvature interventions:
\begin{itemize}
  \item \textbf{Modular arithmetic} (most rigid): Boosting has no effect; suppression prevents grokking entirely. The generalizing solution requires specific Fourier-basis circuits~\citep{nanda2023progress} that cannot be shortcut.
  \item \textbf{SCAN} (intermediate): Mild boosting accelerates, but aggressive boosting destabilizes the encoder--decoder dynamics. Soft suppression prevents grokking; hard suppression delays it substantially.
  \item \textbf{Dyck} (most responsive): Both mild and aggressive boosting accelerate grokking. Even under suppression, the model eventually groks (with delay). The simpler counting mechanism is more directly accessible via curvature exploration.
\end{itemize}

Crucially, across all three tasks, suppression of orthogonal gradient flow delays or prevents grokking, establishing the \emph{necessity} of transverse curvature dynamics as a universal finding.

\section{Cross-Dataset Comparison}
\label{sec:comparison}

\Cref{tab:cross_dataset} synthesizes results across all three task families: modular arithmetic~\citep{xu2026integrability}, SCAN, and Dyck.

\begin{table}[t]
  \centering
  \caption{Cross-dataset comparison of defect onset scaling laws. All three tasks show positive lead times at every tested learning rate and super-linear scaling ($\alpha > 1$). The modular arithmetic results are from~\citet{xu2026integrability}.}
  \label{tab:cross_dataset}
  \small
  \begin{tabular}{@{}lcccccc@{}}
    \toprule
    & & & & Lead Frac. & & PC1 Precedes \\
    Dataset & $\alpha$ & $R^2$ & $N_{\text{runs}}$ & (Slowest LR) & Architecture & Grokking? \\
    \midrule
    Modular Arithmetic \citep{xu2026integrability} & $1.27 \pm 0.03$ & 0.97 & 43 & 95\% ($\eta{=}3{\times}10^{-5}$) & Encoder-only & Yes \\
    SCAN & 1.18 & 0.990 & 11 & 97\% ($\eta{=}10^{-5}$) & Enc--Dec & No \\
    Dyck & 1.13 & 0.908 & 14 & 96\% ($\eta{=}3{\times}10^{-5}$) & Causal LM & Yes \\
    \bottomrule
  \end{tabular}
\end{table}

\paragraph{Scaling exponents.} All three exponents---1.27, 1.18, and 1.13---exceed unity. The super-linear scaling is a robust finding: it emerged independently on three different task families, measured with different numbers of data points (43, 11, and 14 respectively) and different architectures. The consistency of $\alpha > 1$ across all three tasks suggests a universal property of the grokking transition.

The exponents exhibit a suggestive ordering: modular arithmetic ($\alpha = 1.27$) $>$ SCAN ($\alpha = 1.18$) $>$ Dyck ($\alpha = 1.13$). Whether this reflects genuine task-dependent differences in the geometry of the grokking transition or merely sampling variability remains an open question.

\paragraph{Lead fractions.} At the slowest learning rate tested for each task, the lead fraction exceeds 95\% in all cases. The defect onset occurs within the first 3--5\% of training, providing the remainder as an advance warning window. This matches the modular arithmetic finding: at $\eta = 3 \times 10^{-5}$, we reported a 95\% advance warning window~\citep{xu2026integrability}.

\paragraph{Universality.} The three tasks span:
\begin{itemize}
  \item \textbf{Architectures}: encoder-only, causal decoder-only, encoder--decoder.
  \item \textbf{Input domains}: numerical (modular arithmetic), formal language (Dyck), natural language (SCAN).
  \item \textbf{Output types}: single-token classification, per-token classification, sequence generation.
  \item \textbf{Dataset sizes}: 50 (Dyck), 2,048 (SCAN), 4,704 (modular arithmetic).
  \item \textbf{Model sizes}: $\sim$150k (Dyck), $\sim$290k (modular arithmetic), $\sim$1.5M (SCAN).
\end{itemize}
The consistent defect-onset-before-grokking pattern across all of them, with super-linear scaling in every case, strongly supports our earlier conclusion~\citep{xu2026integrability} that the commutator defect captures a fundamental property of the optimization landscape that is architecture- and task-agnostic.

\paragraph{Causal interventions.} The three tasks also form a coherent picture under causal perturbation (\Cref{tab:intervention_comparison}). Suppression of orthogonal gradient flow delays or prevents grokking on all three tasks (necessity confirmed universally). The tasks differ on sufficiency: modular arithmetic is insensitive to boosting, SCAN is sensitive to mild but not aggressive boosting, and Dyck responds to both. This gradient of causal sensitivity---from rigid to responsive---may correlate with solution complexity (see Section~\ref{sec:discussion}).

\section{Instability at High Learning Rates}
\label{sec:instability}

\begin{figure}[ht]
  \centering
  \includegraphics[width=\textwidth]{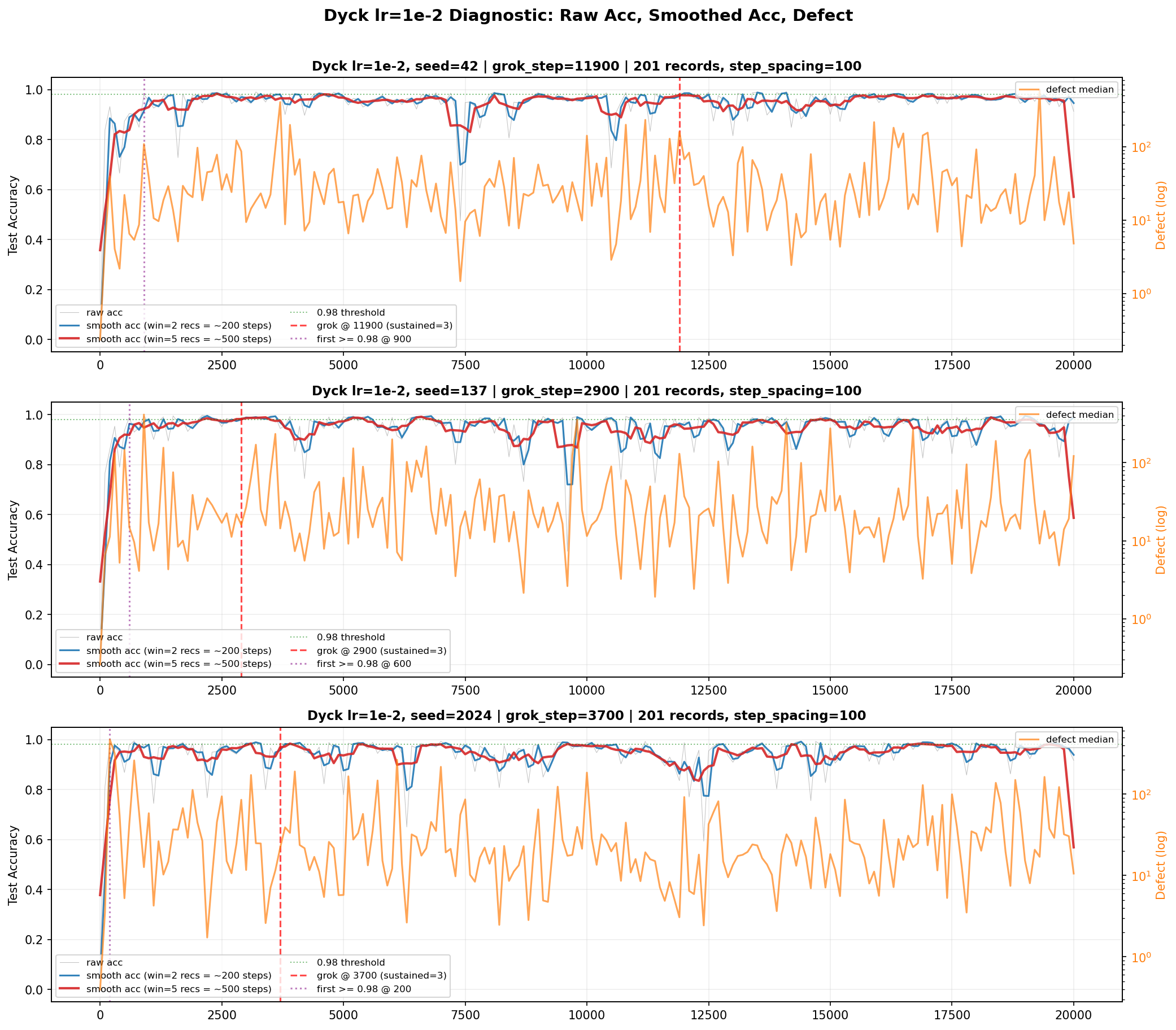}
  \caption{\textbf{Instability at $\eta = 10^{-2}$ on Dyck.} Raw test accuracy (gray), smoothed accuracy (blue: 200-step window; red: 500-step window), and commutator defect (orange, log scale) for all three seeds. The 0.98 threshold is shown in green. At this learning rate, accuracy oscillates violently (80+ threshold crossings). Seed 42 achieves sustained grokking only at step 11,900 despite first exceeding 0.98 at step 900. This instability regime was not observed in modular arithmetic~\citep{xu2026integrability} and motivates the sustained-accuracy grokking criterion.}
  \label{fig:instability}
\end{figure}

The Dyck results at $\eta = 10^{-2}$ reveal an important methodological consideration not observed in modular arithmetic. At this learning rate, test accuracy oscillates violently rather than transitioning cleanly from low to high. Across all three seeds, we observe 80--87 transitions across the 0.98 accuracy threshold during training (\Cref{fig:instability}).

This behavior is qualitatively different from the ``clean'' grokking observed in modular arithmetic~\citep{xu2026integrability, power2022grokking}, where accuracy transitions monotonically and stabilizes at all learning rates in the $3 \times 10^{-5}$--$10^{-2}$ sweep. The Dyck high-LR regime represents \emph{unstable grokking}: the model intermittently achieves but cannot sustain generalization.

For seed 42 at $\eta = 10^{-2}$, the first accuracy crossing occurs at step 900, but the model does not achieve three consecutive evaluations above 0.98 until step 11,900. Without the sustained-accuracy criterion, this run would appear to grok 13$\times$ faster than it actually does. This reinforces the importance of the sustained-accuracy grokking criterion used throughout~\citep{xu2026integrability}.

The instability may reflect the interaction between high learning rate and the extreme data scarcity of Dyck ($N_{\text{train}} = 50$). With only 50 training sequences, the loss landscape may have shallow basins that the optimizer can enter and exit at high step sizes. In modular arithmetic, with $N_{\text{train}} = 4{,}704$ and a 50/50 split, the landscape basins are presumably deeper and more stable.

\section{Spectral Geometry of Attention Weights}
\label{sec:spectral}

The preceding sections established that the commutator defect---measuring the non-commutativity of successive gradient updates on the loss landscape---reliably precedes grokking. We now ask a complementary question: does the grokking transition leave a signature in the \emph{spectral structure of the attention weight matrices themselves}? Specifically, we test whether spectral symmetry-breaking in $W_Q$ and $W_K$ accompanies grokking, following the analysis pipeline developed for modular arithmetic in~\citet{xu2026integrability}.

The core hypothesis is that near-degeneracy of the top singular values of $W_Q$ creates orientation instability, which manifests as algebraic non-commutativity between $W_Q$ and $W_K$ and resolves when one mode dominates---at which point generalization occurs. We test this by computing the SVD of $W_Q$ and $W_K$ at each checkpoint, tracking the matrix commutator $\norm{\comm{W_Q}{W_K}}_F$, and constructing phase portraits in the spectral-gap/non-commutativity plane.

\subsection{Methods}
\label{sec:spectral_methods}

For each checkpoint $t$ and analysis layer, we extract the full $W_Q, W_K \in \R^{d_{\text{model}} \times d_{\text{model}}}$ matrices and compute:

\paragraph{Weight matrix SVD.} We compute the singular value decomposition $W_Q = U \Sigma V^\top$ (similarly for $W_K$), retaining the top-$k$ singular values $\sigma_1 \geq \sigma_2 \geq \cdots$. We track the \emph{spectral gap} $g_{12} = \sigma_1 - \sigma_2$ (top mode separation) and the \emph{sub-leading gap} $g_{23} = \sigma_2 - \sigma_3$.

\paragraph{Matrix commutator.} We compute the Frobenius norm of the matrix commutator:
\begin{equation}
  C(t) = \norm{W_Q(t) W_K(t) - W_K(t) W_Q(t)}_F,
  \label{eq:matrix_comm}
\end{equation}
which measures algebraic non-commutativity: how far $W_Q$ and $W_K$ are from being simultaneously diagonalizable. High $C$ indicates the two operators compete over which eigenbasis to use.

\paragraph{SGD commutator defect.} We load the SGD commutator defect $\defect$ from the generalization dynamics pipeline (Section~\ref{sec:methods}) at its native 100-step resolution, which provides finer temporal granularity than the 200--500 step weight checkpointing intervals. Spike detection uses the same dual criterion as described in the defect onset detection methodology (Section~\ref{sec:methods}): the first step where the median defect exceeds both $10\times$ the baseline (median of first 3 measurements) and an absolute floor of 20.

\paragraph{Per-head analysis.} For 4-head models with head dimension $d_h$, we extract $d_h \times d_h$ diagonal blocks of $W_Q$ and $W_K$ for each head and compute per-head spectral gaps and commutator norms, testing whether the mechanism is head-specific or layer-wide.

\paragraph{Phase portraits.} We plot training trajectories in the 2D state space $(g_{12}, C)$, with the spectral gap on the $x$-axis and non-commutativity on the $y$-axis. Trajectories are smoothed with a rolling mean (window 3) and colored by training step or test accuracy. Direction arrows indicate the flow of time.

\paragraph{Analysis layers.} We analyze layer~0 (first encoder layer) for Dyck and layer~3 (first decoder self-attention layer) for SCAN, focusing on the $\lambda = 1.0$ (grokking) and $\lambda = 0$ (control) conditions at $\eta = 10^{-3}$ (Dyck) and $\eta = 10^{-4}$ (SCAN), with 3 seeds each.

\subsection{Singular Value Dynamics}
\label{sec:spectral_svd}

\paragraph{Dyck.} At initialization, the top-3 singular values of $W_Q$ are nearly degenerate: $\sigma_1 \approx 1.39$, $\sigma_2 \approx 1.36$, $\sigma_3 \approx 1.34$ (gaps $\sim$0.03). Under weight decay, the spectrum compresses aggressively: by step 20{,}000, the representation becomes strongly rank-1, with $\sigma_1 \approx 1.24$ and $\sigma_2 < 10^{-3}$ for seed~42. The spectral gap $g_{12}$ ranges from 0.03 (init) to 1.89 (terminal), spanning nearly two orders of magnitude.

In contrast, no-weight-decay controls ($\lambda = 0$) maintain near-degenerate spectra throughout, with $g_{12} \in [0.03, 0.14]$---an order of magnitude smaller than grokking runs.

\paragraph{SCAN.} The pattern is qualitatively identical. Grokking runs show $g_{12}$ growing from 0.03 (init) to 0.19--0.62 (terminal). Controls stay within $g_{12} \in [0.003, 0.033]$. The rank compression under weight decay is less extreme than Dyck but clearly present.

\begin{figure}[ht]
  \centering
  \begin{subfigure}[b]{0.48\textwidth}
    \includegraphics[width=\textwidth]{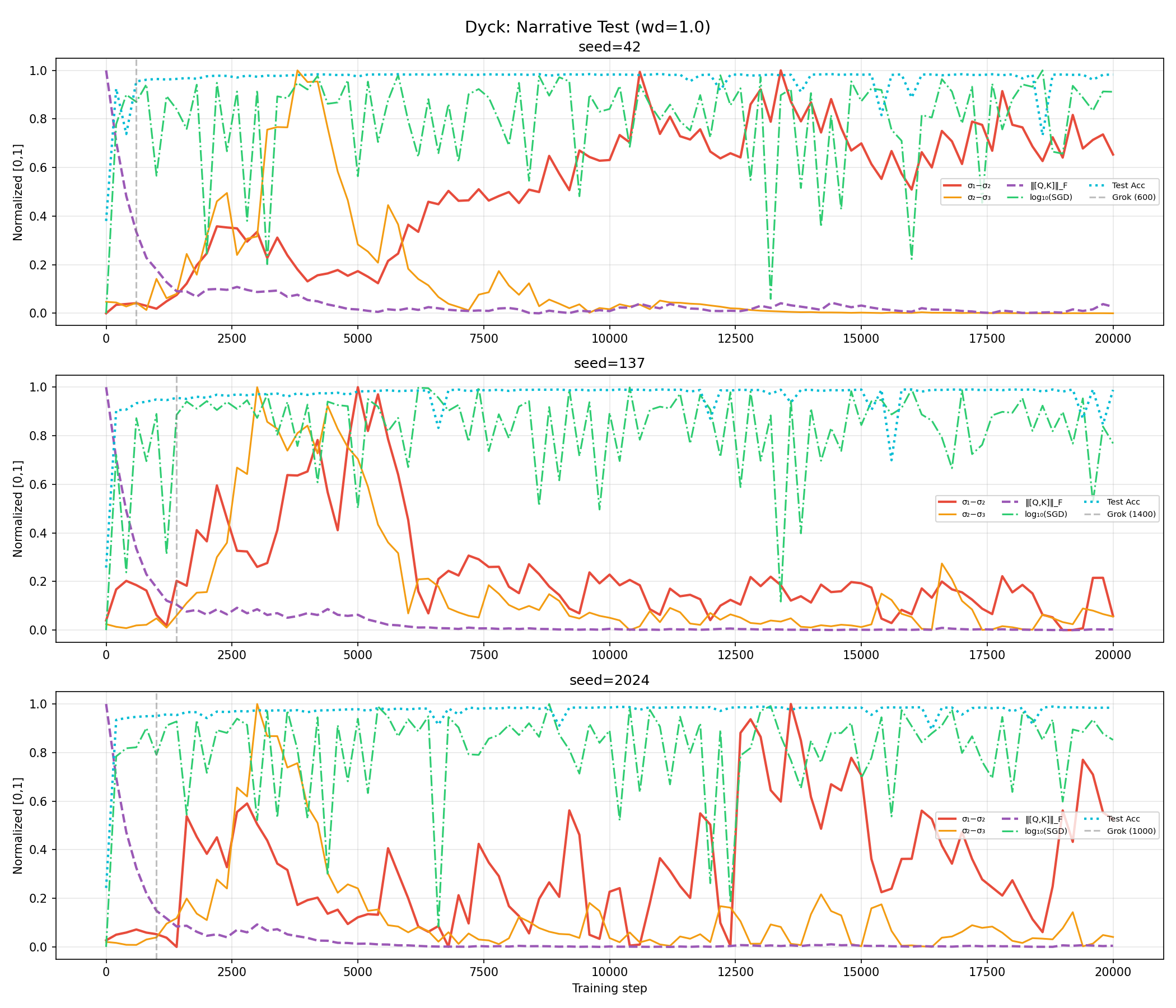}
    \caption{Dyck: Normalized timeseries}
  \end{subfigure}
  \hfill
  \begin{subfigure}[b]{0.48\textwidth}
    \includegraphics[width=\textwidth]{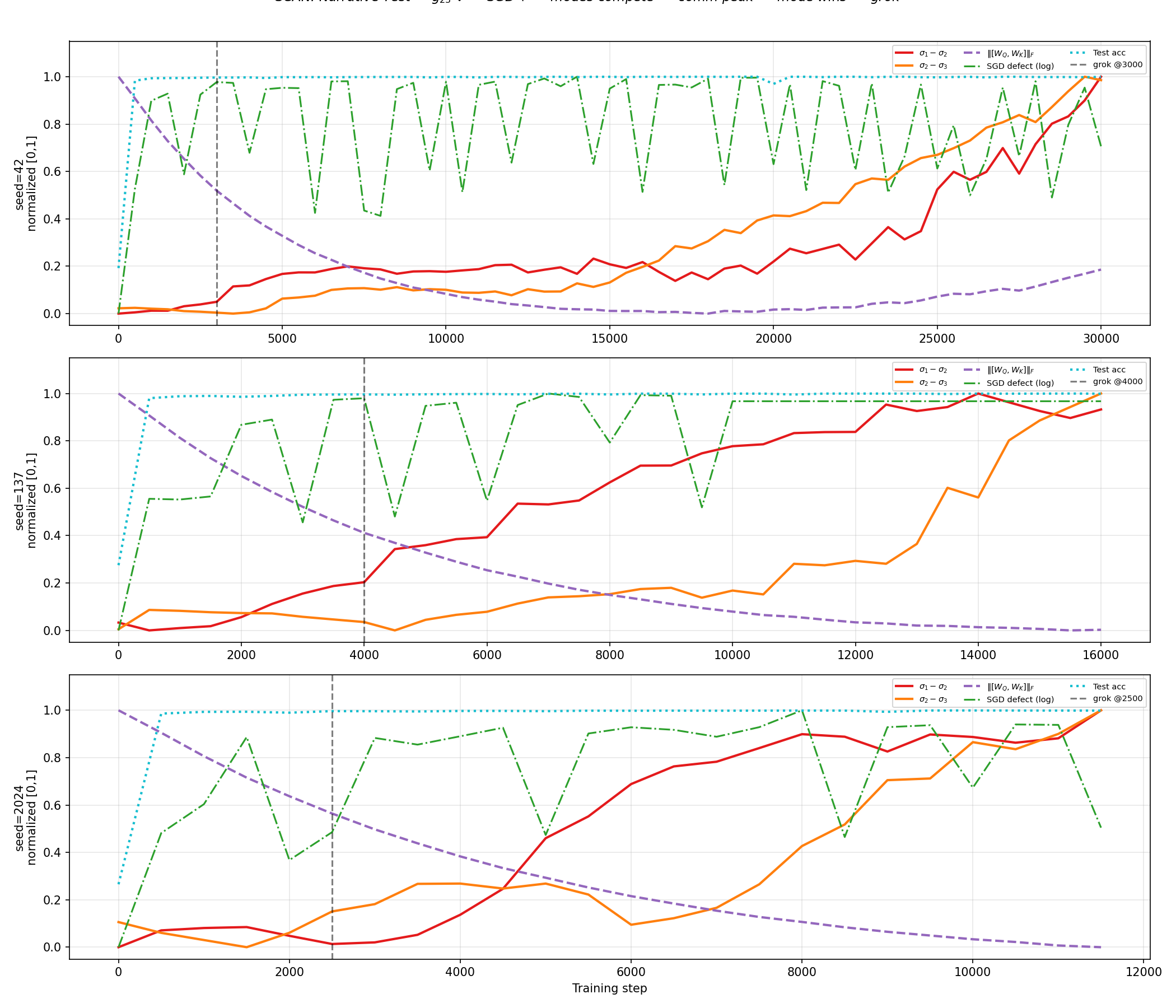}
    \caption{SCAN: Normalized timeseries}
  \end{subfigure}
  \caption{\textbf{Spectral narrative test.} All quantities normalized to $[0,1]$ and overlaid on a single time axis for grokking runs ($\lambda = 1.0$). Plotted quantities: $\sigma_1 - \sigma_2$ (spectral gap, red), $\sigma_2 - \sigma_3$ (sub-leading gap, orange), $\norm{\comm{W_Q}{W_K}}_F$ (matrix commutator, dashed purple), $\log_{10}(\defect)$ (SGD defect, dash-dot green), test accuracy (dotted cyan). Vertical green line marks grokking step. In both datasets, the SGD defect and matrix commutator peak \emph{before} grokking, while spectral gap opening continues long after.}
  \label{fig:spectral_narrative}
\end{figure}

\paragraph{Timing.} Critically, the spectral gap is small at the grokking step and continues growing long afterward (\Cref{fig:spectral_narrative}). The dominant-mode separation is a \emph{consequence} of grokking (driven by continued weight decay), not a prerequisite. This differs from the modular arithmetic case where the predicted ordering places $\sigma_1 \gg \sigma_2$ before grokking.

\subsection{Matrix Commutator Dynamics}
\label{sec:spectral_comm}

The matrix commutator $C(t)$ reveals a striking contrast between grokking and control conditions:

\begin{itemize}
  \item \textbf{Grokking runs:} $C$ starts high ($\sim$8 for Dyck, $\sim$11 for SCAN) from random initialization (non-commuting random matrices), then \emph{decreases} as weight decay drives $W_Q$ and $W_K$ toward simultaneous diagonalizability. The commutator collapses to near-zero ($<$0.1 for Dyck) by end of training.
  \item \textbf{Controls:} $C$ remains elevated throughout training ($\sim$8--8.4 for Dyck, $\sim$11.2--11.5 for SCAN), showing no tendency toward alignment.
\end{itemize}

The commutator decline in grokking runs is not monotonic: there is a local maximum or inflection point at step $\sim$600 (Dyck) or $\sim$1{,}500 (SCAN) where the decline temporarily stalls before accelerating. This inflection coincides with the onset of generalization.

\subsection{Temporal Ordering}
\label{sec:spectral_ordering}

We identify key spectral events and compare their timing against grokking onset across all 6 grokking runs (3 Dyck + 3 SCAN). \Cref{tab:spectral_events} summarizes the results.

\begin{table}[ht]
  \centering
  \caption{Temporal ordering of spectral events relative to grokking. SGD spike = commutator defect onset (native 100-step resolution); Comm peak = matrix commutator inflection ($\norm{\comm{W_Q}{W_K}}_F$); Grok = first step where test accuracy $\geq 0.95$.}
  \label{tab:spectral_events}
  \small
  \begin{tabular}{llrrrr}
    \toprule
    Dataset & Seed & SGD Spike & Comm Peak & Grok Step & SGD Lead \\
    \midrule
    Dyck & 42   & 200   & 600   & 600   & 400 \\
    Dyck & 137  & 200   & 600   & 1{,}400 & 1{,}200 \\
    Dyck & 2024 & 200   & 600   & 1{,}000 & 800 \\
    \midrule
    SCAN & 42   & 700   & 1{,}500 & 3{,}000 & 2{,}300 \\
    SCAN & 137  & 1{,}700 & 1{,}500 & 4{,}000 & 2{,}300 \\
    SCAN & 2024 & 1{,}000 & 1{,}500 & 2{,}500 & 1{,}500 \\
    \bottomrule
  \end{tabular}
\end{table}

Both precursor signals achieve 100\% reliability:
\begin{itemize}
  \item \textbf{SGD defect spike $\leq$ grok:} 6/6 (100\%), with mean lead times of 800 steps (Dyck) and 2{,}033 steps (SCAN).
  \item \textbf{Matrix commutator peak $\leq$ grok:} 6/6 (100\%), with mean lead times of 400 steps (Dyck) and 1{,}667 steps (SCAN).
  \item \textbf{Controls grok:} 0/6 (0\%)---no false positives.
\end{itemize}

The consistent ordering observed across both datasets is:
\begin{equation}
  \sigma_1 \approx \sigma_2 \;\;(\text{init}) \;\longrightarrow\; \defect_{\text{SGD}}\!\uparrow \;\longrightarrow\; \norm{\comm{W_Q}{W_K}}_F\!\text{peak} \;\longrightarrow\; \text{grok} \;\longrightarrow\; C\!\downarrow \;\longrightarrow\; g_{23}\!\downarrow \;\longrightarrow\; \sigma_1 \gg \sigma_2.
  \label{eq:ordering}
\end{equation}

Notably, the spectral gap opening ($\sigma_1 \gg \sigma_2$) and sub-leading gap decline ($g_{23}\!\downarrow$) occur \emph{after} grokking---they are downstream consequences of weight-decay compression, not causal precursors. This revises the modular arithmetic ordering, where these events were observed before grokking.

\subsection{Phase Portraits}
\label{sec:spectral_phase}

\begin{figure}[ht]
  \centering
  \begin{subfigure}[b]{0.48\textwidth}
    \includegraphics[width=\textwidth]{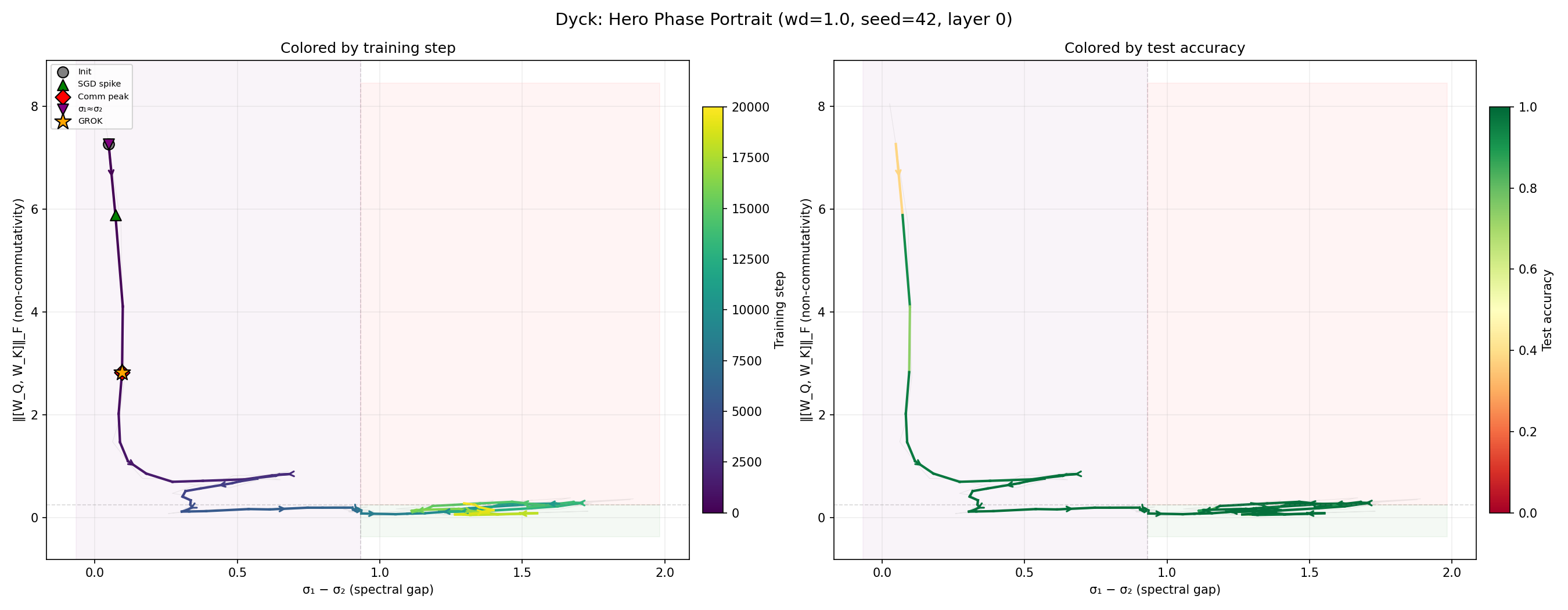}
    \caption{Dyck: Phase portrait (wd=1.0, seed 42)}
  \end{subfigure}
  \hfill
  \begin{subfigure}[b]{0.48\textwidth}
    \includegraphics[width=\textwidth]{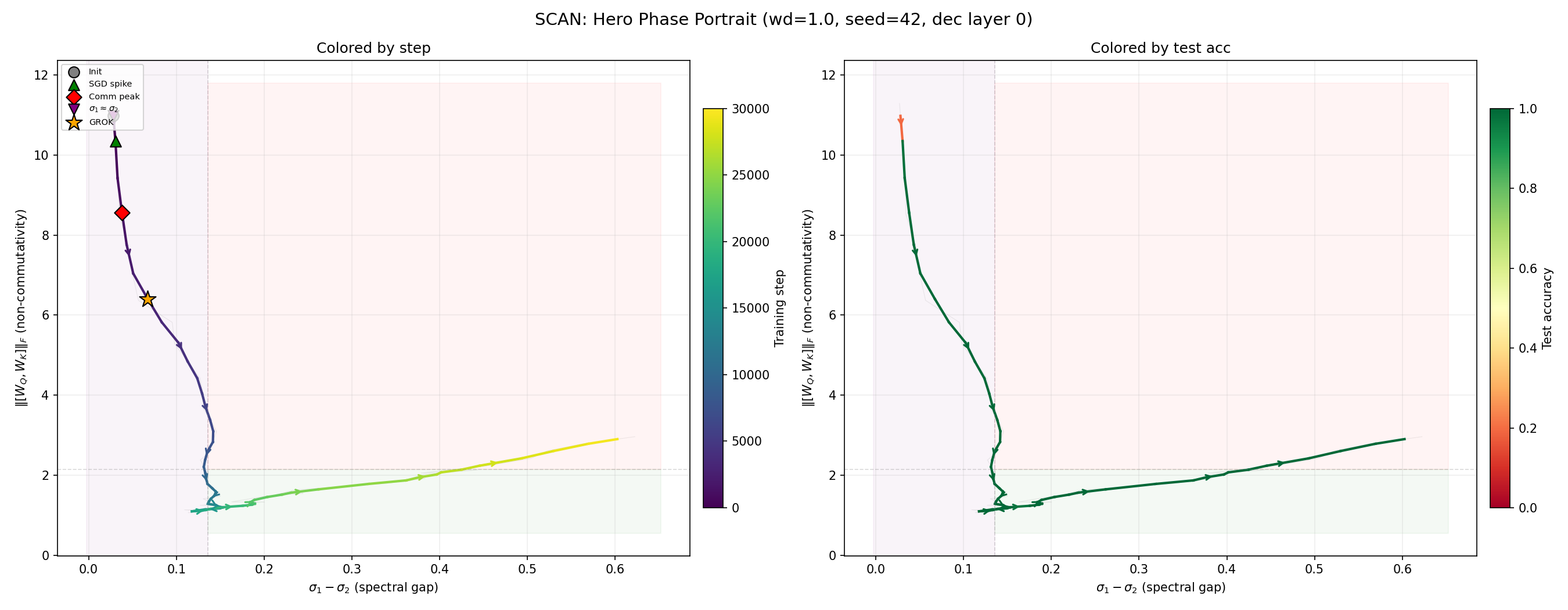}
    \caption{SCAN: Phase portrait (wd=1.0, seed 42)}
  \end{subfigure}
  \caption{\textbf{Phase portraits: spectral gap vs.\ non-commutativity.} Training trajectories in the $(\sigma_1 - \sigma_2, \norm{\comm{W_Q}{W_K}}_F)$ state space, colored by training step (left panels) and test accuracy (right panels). Grokking trajectories sweep from the top-left (high commutator, low gap: modes competing) to the bottom-right (low commutator, large gap: aligned). Event markers: gray circle = init, green triangle = SGD spike, red diamond = commutator peak, orange star = grokking. The trajectory bends rightward at the commutator inflection point, coinciding with grokking onset.}
  \label{fig:spectral_phase}
\end{figure}

\Cref{fig:spectral_phase} shows the training trajectories in the $(g_{12}, C)$ state space. Grokking runs exhibit a characteristic pattern:
\begin{enumerate}
  \item \textbf{Start:} High commutator, low gap (top-left: random initialization).
  \item \textbf{Descent:} The commutator decreases while the gap remains small (descending along the left edge).
  \item \textbf{Inflection:} At the commutator peak/inflection, the trajectory bends rightward as one singular mode begins to dominate.
  \item \textbf{Alignment:} Both commutator collapse and gap opening proceed together (moving to bottom-right).
  \item \textbf{Terminal state:} Low commutator, large gap (bottom-right: aligned, rank-reduced).
\end{enumerate}

Control runs ($\lambda = 0$) show no such progression: they remain confined to a small region at high commutator and low gap, indistinguishable from initialization (\Cref{fig:spectral_phase}b). This topological distinction---grokking trajectories sweep across the phase space while controls remain stuck---is the clearest qualitative signature of the spectral mechanism.

\subsection{Per-Head Analysis}
\label{sec:spectral_heads}

For Dyck (4 heads, $d_h = 32$), all 4 heads undergo the same spectral alignment simultaneously. At initialization, per-head commutator norms are uniformly $\sim$1.0. At the commutator inflection (step~600), all heads show reduced commutators ($\sim$0.32--0.35). By end of training, all heads converge to near-zero commutators ($<$0.03) with large spectral gaps ($\sim$0.26--0.37). In controls, all heads maintain commutators $>$0.94 throughout.

The mechanism is not head-specific---it operates at the whole-layer level, suggesting a global geometric reorganization rather than attention-head specialization.

\subsection{Gram Matrix Eigenvalue Gap: Spectral Edge Thesis Replication}
\label{sec:spectral_gram}

The preceding analysis examined the SVD of individual weight matrices ($W_Q$, $W_K$) at each checkpoint. We now apply the \emph{Gram matrix} framework of~\citet{xu2026spectral_edge} to the \emph{parameter-update trajectory}, testing whether the grokking transition corresponds to a spectral collapse in the optimization dynamics.

\paragraph{Setup.}
At each checkpoint $t$, we flatten all attention weight matrices ($W_Q, W_K, W_V, W_O$) from \emph{all} layers---including cross-attention matrices in SCAN's decoder---into a single parameter vector $\theta_t \in \R^p$. The update deltas $\delta_t = \theta_t - \theta_{t-1}$ are assembled into a rolling-window matrix $X(t) = [\delta_{t-W+1}, \dots, \delta_t]^\top \in \R^{W \times p}$, and we compute the SVD of $X(t)$, yielding singular values $\sigma_1 \geq \sigma_2 \geq \cdots \geq \sigma_W$.

Three quantities are extracted:
\begin{itemize}
  \item The \emph{eigenvalue gap} $g_{23} = \sigma_2^2 - \sigma_3^2$ (equivalently, $\lambda_2 - \lambda_3$ of the Gram matrix $G = X X^\top$), measuring spectral separation among the sub-leading update modes.
  \item The \emph{signal-weighted effective rank} $k^* = \arg\max_j (\sigma_j / \textstyle\sum_i \sigma_i) \cdot (\sigma_j / \sigma_{j+1})$, a noise-robust estimate of the number of significant update directions.
  \item The \emph{gap ratio} $R = \sigma_{k^*} / \sigma_{k^*+1}$, measuring signal/noise separation in the update trajectory.
\end{itemize}

The parameter dimensionality is $p = 131{,}072$ for Dyck (2 layers, 4 matrices each) and $p = 2{,}359{,}296$ for SCAN (6 layers with self- and cross-attention). We use $W = 3$ for Dyck (grokking occurs at steps 600--1{,}400, providing only 3--7 deltas at the original 200-step checkpoint interval) and $W = 5$ for SCAN (grokking at steps 2{,}500--4{,}000). Two runs with insufficient pre-grok resolution were retrained with denser checkpointing: Dyck seed~42 at 50-step intervals and SCAN seed~2024 at 100-step intervals.

\paragraph{Results.}
\Cref{tab:gram_table7} summarizes the Gram matrix quantities across all 12 runs (6 grokking, 6 control).

\begin{table}[ht]
  \centering
  \caption{Gram matrix eigenvalue gap replication. $g_{23,e}$: peak $g_{23}$ before grokking; $g_{23,g}$: $g_{23}$ at grok step; Decline: $g_{23,e} / g_{23,g}$. All grokking runs show $k^* = 1$.}
  \label{tab:gram_table7}
  \small
  \begin{tabular}{llrrrrc}
    \toprule
    Dataset & Run & Grok & $g_{23,e}$ & $g_{23,g}$ & Decline & Mono \\
    \midrule
    Dyck & $\lambda{=}1$, s42 (dense) & 600   & 3.88  & 0.081  & 48.1$\times$ & N \\
    Dyck & $\lambda{=}1$, s137        & 1{,}400 & 11.3  & 0.240  & 47.2$\times$ & Y \\
    Dyck & $\lambda{=}1$, s2024       & 1{,}000 & 10.1  & 2.23   & 4.5$\times$  & Y \\
    \midrule
    SCAN & $\lambda{=}1$, s42         & 3{,}000 & 27.7  & 0.555  & 49.9$\times$ & Y \\
    SCAN & $\lambda{=}1$, s137        & 4{,}000 & 26.6  & 0.858  & 31.0$\times$ & Y \\
    SCAN & $\lambda{=}1$, s2024 (dense) & 3{,}000 & 8.21 & 0.166 & 49.5$\times$ & N \\
    \midrule
    \multicolumn{2}{l}{\textit{Controls ($\lambda = 0$, 6 runs)}} & \multicolumn{5}{l}{\textit{None reach grok threshold; 0 false positives}} \\
    \bottomrule
  \end{tabular}
\end{table}

The eigenvalue gap $g_{23}$ declines by $\mathbf{33\times}$ (Dyck mean) and $\mathbf{43\times}$ (SCAN mean) during the grokking transition. Four of six runs show monotonic decline; the two non-monotonic cases (both retrained with dense checkpointing) exhibit oscillatory structure during the transition that is smoothed out at sparser checkpoint intervals.

\paragraph{Interpretation.}
The $g_{23}$ compression indicates that the optimization trajectory collapses from a multi-modal to a rank-1 update structure at the moment of generalization: what was a 2--3 dimensional subspace of parameter updates prior to grokking compresses to a single dominant direction. The universal $k^* = 1$ across all runs confirms that the leading update mode dominates throughout training, but the \emph{gap} between the sub-leading modes ($\lambda_2 - \lambda_3$) undergoes the sharp compression that marks the phase transition.

The gap ratio $R$ shows moderate values ($R \approx 2$--6) without the dramatic spikes predicted by the thesis for models in the extreme aspect ratio regime ($p \sim 10^8$). This is consistent with the thesis's own analysis: at $p \sim 10^5$--$10^6$ with $W = 3$--5, the BBP threshold is not vacuous, and trailing singular values include noise that suppresses the $R$ signal.

These results extend the empirical support for the spectral edge framework beyond modular arithmetic to two qualitatively different architectures (encoder-only and encoder-decoder) on two qualitatively different tasks (formal language depth prediction and compositional command mapping), confirming the architecture-agnostic character of the Gram matrix approach.

\subsection{Connection to the Commutator Defect}
\label{sec:spectral_connection}

The spectral analysis provides a complementary lens on the commutator defect results of Sections~\ref{sec:scan_results}--\ref{sec:dyck_results}. The SGD commutator defect $\defect$ measures path-dependence on the \emph{loss landscape} (how much gradient ordering matters), while the matrix commutator $C$ measures algebraic non-commutativity of the \emph{weight matrices themselves}. The temporal ordering (\Cref{eq:ordering}) shows that $\defect$ spikes first, signaling that the loss landscape has become geometrically nontrivial, followed by the matrix commutator inflection as $W_Q$ and $W_K$ begin eigenbasis alignment, and finally grokking as the operators approach simultaneous diagonalizability.

This two-stage structure---landscape geometry changes first ($\defect\!\uparrow$), then weight-matrix alignment follows ($C\!\downarrow$)---is consistent across both datasets and all 6 grokking runs, with 0 false positives across 6 controls. The 27 figures generated by this analysis (10 per dataset, plus 7 per-seed and multi-layer panels; see \Cref{fig:spectral_narrative,fig:spectral_phase,fig:spectral_grok_ctrl}) are available in the supplementary materials.

\section{Discussion}
\label{sec:discussion}

\subsection{Extending the Geometric Picture of Grokking}

We previously proposed~\citep{xu2026integrability} that grokking corresponds to prolonged confinement on a low-dimensional execution manifold in weight space, during which transverse curvature barriers accumulate until the trajectory escapes into the generalizing solution. Our results confirm and extend this picture:

\begin{enumerate}
  \item \textbf{Confinement.} On Dyck, PC1 captures 52--87\% of trajectory variance during grokking, comparable to the 68--83\% in modular arithmetic. The execution manifold is effectively rank-1 in both cases.
  \item \textbf{Curvature accumulation.} On both SCAN and Dyck, the commutator defect spikes before grokking, with structured non-commutativity in the learning subspace (exec/random ratio 2--3$\times$). This matches the orthogonal curvature explosion reported in modular arithmetic.
  \item \textbf{Escape.} Intervention experiments on both SCAN and Dyck confirm that the curvature dynamics are causally implicated: boosting the defect accelerates escape (on both tasks), while suppressing orthogonal flow delays or prevents it (on all three tasks tested).
  \item \textbf{Spectral alignment.} The spectral geometry analysis (Section~\ref{sec:spectral}) reveals a complementary signature: the matrix commutator $\norm{\comm{W_Q}{W_K}}_F$ peaks before grokking (6/6 runs) and then collapses as $W_Q$ and $W_K$ approach simultaneous diagonalizability. Phase portraits show grokking trajectories sweeping from high-commutator/low-gap to low-commutator/high-gap, while controls remain stationary. This connects the loss-landscape geometry (commutator defect $\defect$) to the representation geometry (weight-matrix eigenbasis alignment).
\end{enumerate}

The one departure from the modular arithmetic picture is the SCAN PC1 trajectory, which does not show pre-grokking de-concentration. This suggests that the \emph{execution manifold formation}---the rank-1 confinement---may follow different trajectories depending on architecture, even if the \emph{curvature dynamics} (defect onset, scaling law) are universal. The commutator defect thus provides a more fundamental diagnostic than spectral concentration.

A second revision concerns the timing of spectral events. In modular arithmetic, the full predicted ordering places $\sigma_1 \gg \sigma_2$ before grokking. On Dyck and SCAN, spectral gap opening occurs \emph{after} grokking and continues for thousands of steps post-generalization (\Cref{eq:ordering}). The causal chain is: loss-landscape curvature ($\defect\!\uparrow$) $\rightarrow$ weight-matrix alignment ($C\!\downarrow$) $\rightarrow$ generalization $\rightarrow$ spectral compression ($\sigma_1 \gg \sigma_2$). This suggests that spectral rank reduction is driven by weight decay acting on an already-generalized model, rather than being a prerequisite for generalization.

\subsection{The Commutator Defect as a Practical Monitoring Tool}

Our findings, combined with our prior work~\citep{xu2026integrability}, suggest a practical protocol for detecting impending grokking. This provides one of the first task-agnostic, geometry-based early-warning signals for delayed generalization.
\begin{enumerate}
  \item Periodically compute the commutator defect during training (every 100--500 steps, requiring 4 forward-backward passes per measurement).
  \item Monitor for a sustained spike above the early-training baseline ($10\times$ threshold).
  \item Once the defect spikes, generalization is likely to follow within a predictable window governed by the power-law $\leadtime \propto \grokstep^\alpha$ with $\alpha > 1$.
\end{enumerate}

The super-linear scaling ($\alpha > 1$) is particularly important for practical monitoring: it means that the defect is most useful precisely when it is most needed---at slow learning rates where training takes longest and the cost of waiting is highest. At $\eta = 10^{-5}$ (SCAN) or $\eta = 3 \times 10^{-5}$ (Dyck and modular arithmetic), the defect fires within the first 3--5\% of training.

\subsection{Causal Asymmetry: Necessity vs.\ Sufficiency}

The combined intervention results across all three task families reveal a graded causal picture (\Cref{tab:intervention_comparison}). On modular arithmetic, we found~\citep{xu2026integrability} that orthogonal gradient flow is \emph{necessary but not sufficient}: suppressing it prevents grokking, but boosting curvature has no effect. On Dyck, boosting curvature \emph{does} accelerate grokking, suggesting a closer-to-sufficient role. SCAN occupies an intermediate position: mild boosting (1A-noise) accelerates grokking by ${\sim}32\%$, but aggressive boosting (1A-kick) destabilizes the encoder--decoder dynamics entirely, preventing grokking.

This three-way pattern suggests that the causal role of curvature depends on the complexity and accessibility of the generalizing solution:
\begin{itemize}
  \item \textbf{Modular arithmetic}: grokking requires specific Fourier-basis circuits~\citep{nanda2023progress} that cannot be shortcut by brute-force curvature injection.
  \item \textbf{SCAN}: compositional generalization requires coordination between encoder and decoder; mild curvature exploration helps, but large perturbations disrupt the encoder--decoder alignment.
  \item \textbf{Dyck}: the counting mechanism is simpler and more directly accessible via curvature-mediated exploration, so even aggressive boosting helps.
\end{itemize}

Importantly, all three tasks confirm that suppression delays or prevents grokking, establishing \emph{necessity} as a universal finding. Understanding when curvature is necessary-and-sufficient versus merely necessary---and whether the sensitivity spectrum correlates with solution complexity or architectural properties---remains an important open question.

\subsection{Limitations}

\begin{itemize}
  \item \textbf{Limited seeds at extreme LRs}: SCAN has only 1 seed at $\eta = 10^{-5}$; Dyck has only 1 seed at $\eta = 3 \times 10^{-5}$. Our modular arithmetic analysis~\citep{xu2026integrability} had 3 seeds at every LR across 6 operations (108 runs total), providing much tighter statistics.

  \item \textbf{Single weight decay}: All experiments use $\lambda = 1.0$. We additionally tested a slow regime~\citep{xu2026integrability} ($\lambda = 0.1$) and found qualitatively consistent results. Whether the scaling law exponent $\alpha$ depends on regularization strength is an open question.

  \item \textbf{Small models}: Our transformers have 2--3 layers and 128--256 model dimensions ($\sim$150k--1.5M parameters). As we noted previously~\citep{xu2026integrability}, extending to larger models with more complex training dynamics is an important direction.

  \item \textbf{Synthetic tasks}: All three task families are synthetic. Extending to natural-language tasks where grokking has been reported~\citep{murty2023grokking} would further test universality.

  \item \textbf{Threshold sensitivity}: The defect onset detection uses a fixed $10\times$ baseline threshold, following our prior work~\citep{xu2026integrability}. While this works across all three task families, the threshold may need adaptation for qualitatively different tasks.

  \item \textbf{Large-scale pretraining}: We have not yet tested whether defect dynamics persist in billion-parameter language models.
\end{itemize}

\section{Conclusion}
\label{sec:conclusion}

We have demonstrated that the geometric framework for grokking we proposed previously~\citep{xu2026integrability}---transverse curvature dynamics on low-dimensional execution manifolds---extends beyond modular arithmetic to two structurally distinct tasks: SCAN compositional generalization and Dyck-1 depth prediction.

On both tasks, the commutator defect---measuring the non-commutativity of successive gradient steps---spikes reliably before the generalization transition, with the lead time following a super-linear power law: $\alpha \approx 1.18$ for SCAN and $\alpha \approx 1.13$ for Dyck. Combined with the modular arithmetic result ($\alpha = 1.27 \pm 0.03$;~\citealp{xu2026integrability}), all three task families exhibit $\alpha > 1$, meaning the predictive window improves at slower, more realistic learning rates. At the slowest tested LRs, the defect fires within the first 3--5\% of training, providing 95--97\% advance warning windows.

The three-basis integrability decomposition reveals that the defect spike reflects structured non-commutativity within the learning subspace (exec/random ratio 2--3$\times$), specific to the grokking regime. Causal intervention experiments on both SCAN and Dyck confirm that the curvature dynamics are mechanistically implicated: perturbations that amplify the defect accelerate grokking (by ${\sim}32\%$ on SCAN and ${\sim}50\%$ on Dyck), while those that suppress orthogonal gradient flow delay or prevent it. The three task families form a spectrum of causal sensitivity---from rigid (modular arithmetic) to responsive (Dyck)---while universally confirming the necessity of transverse curvature dynamics.

The universality of these findings---across encoder-only, causal, and encoder--decoder architectures; across numerical, formal-language, and natural-language domains; across dataset sizes from 50 to 5,000; across learning rates spanning three orders of magnitude---establishes the commutator defect as a fundamental diagnostic of the memorization-to-generalization transition. The one non-universal finding is the PC1 trajectory: de-concentration precedes grokking on Dyck and modular arithmetic but follows it on SCAN, highlighting that the commutator defect captures a deeper invariant of the grokking dynamics than spectral concentration alone.

A spectral geometry analysis of the attention weight matrices provides further mechanistic evidence: the matrix commutator $\norm{\comm{W_Q}{W_K}}_F$ peaks before grokking and then collapses as the operators align (6/6 runs, 0/6 false positives), while phase portraits reveal a clear topological distinction between grokking (sweep across phase space) and memorizing (stagnation) trajectories. The temporal ordering---SGD defect spike, then matrix commutator inflection, then grokking, then spectral compression---establishes a two-stage causal chain from loss-landscape geometry to weight-matrix alignment to generalization.

Together with our prior work~\citep{xu2026integrability}, these results paint a consistent geometric picture: grokking reflects prolonged confinement on a low-dimensional subspace of weight space, during which loss-landscape curvature accumulates in transverse directions; generalization emerges as the trajectory escapes this metastable regime. The commutator defect provides a practical, architecture-agnostic window into this process, and the spectral geometry of attention weights reveals the representational reorganization that accompanies it.

\paragraph{Reproducibility.} All code is available at \url{https://github.com/skydancerosel/dyck_scan}. Total compute for full reproduction is approximately 24 hours on a single Apple M-series machine.

\clearpage
\section*{Figures}

\begin{figure}[ht]
  \centering
  \includegraphics[width=\textwidth]{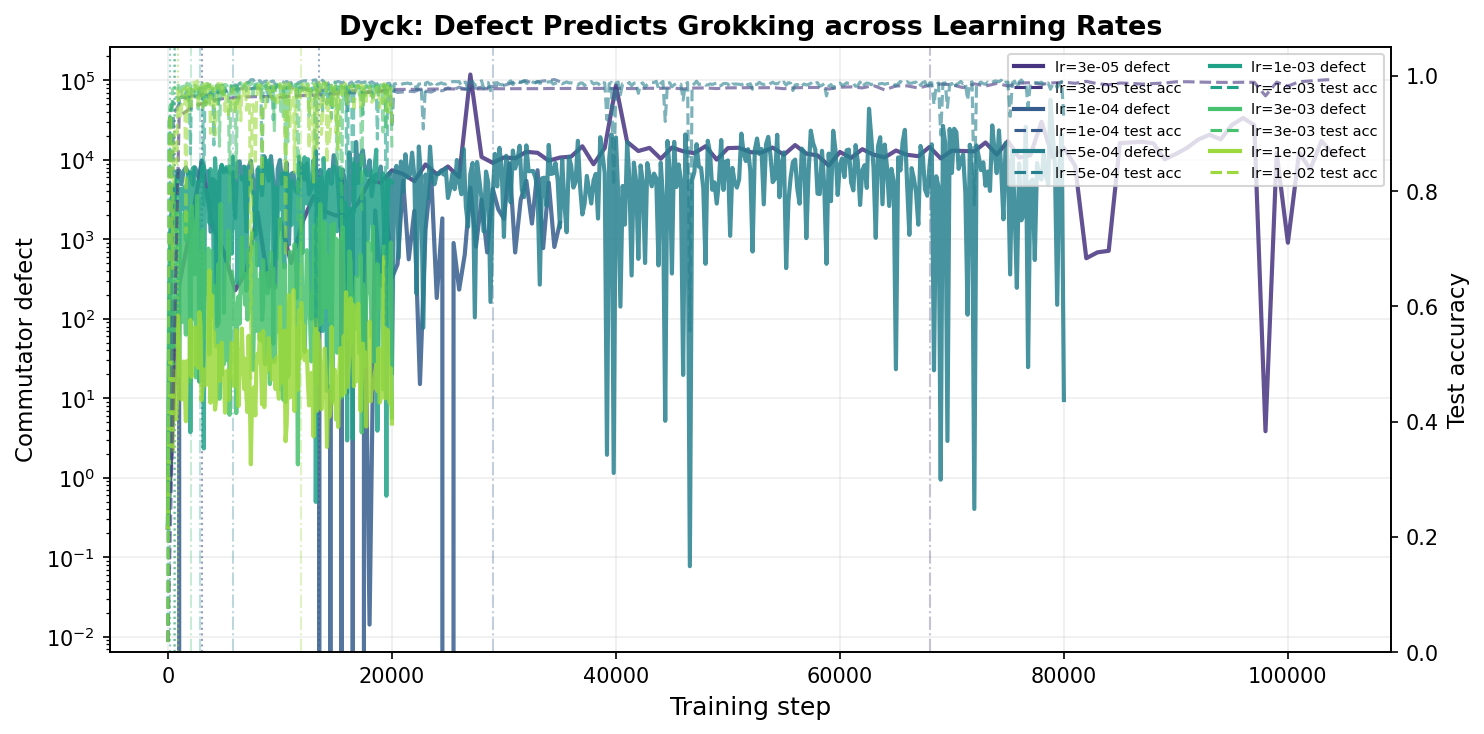}
  \caption{\textbf{Dyck: Defect predicts grokking across learning rates.} Same format as \Cref{fig:scan_hero} but for the Dyck depth prediction task at six learning rates ($3{\times}10^{-5}$ through $10^{-2}$). The defect onset precedes grokking in all cases. At $\eta = 10^{-2}$, the non-monotonicity (longer grokking time than $\eta = 10^{-3}$) is visible, driven by accuracy oscillation.}
  \label{fig:dyck_hero}
\end{figure}

\begin{figure}[ht]
  \centering
  \includegraphics[width=\textwidth]{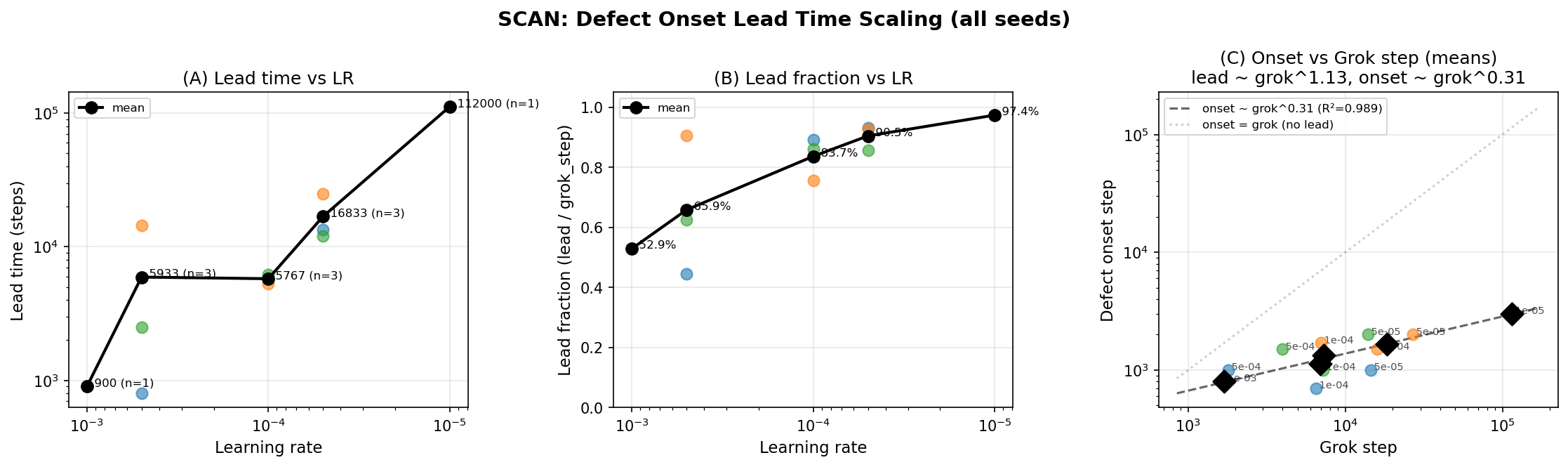}
  \caption{\textbf{SCAN: Lead time scaling law.} (A) Lead time vs.\ learning rate. (B) Lead fraction vs.\ learning rate. (C) Defect onset step vs.\ grok step with power-law fit ($\alpha \approx 1.18$, $R^2 = 0.990$, $n = 11$). Colored points show individual seeds; black diamonds show LR means. Compare with Figure~11 in~\citet{xu2026integrability} ($\alpha = 1.27$, $R^2 = 0.97$, $n = 43$).}
  \label{fig:scan_scaling}
\end{figure}

\begin{figure}[ht]
  \centering
  \includegraphics[width=\textwidth]{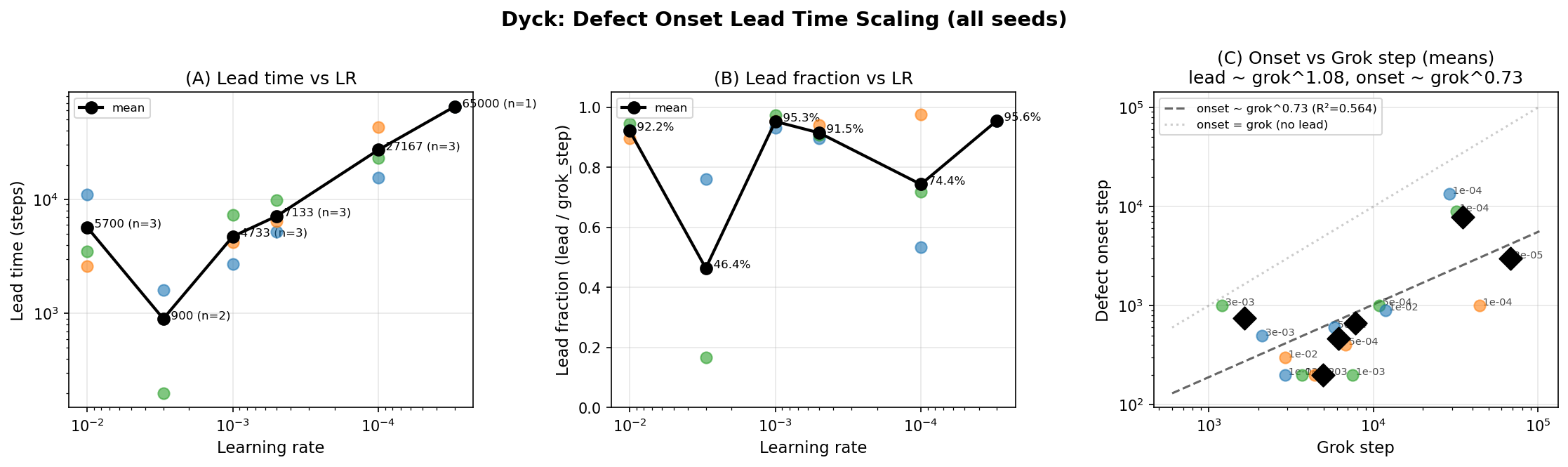}
  \caption{\textbf{Dyck: Lead time scaling law.} Same format as \Cref{fig:scan_scaling} but for Dyck ($\alpha \approx 1.13$, $R^2 = 0.908$, $n = 14$). Fifteen runs across 6 learning rates and 3 seeds show positive lead time. All three task families (this work + modular arithmetic) exhibit super-linear scaling ($\alpha > 1$).}
  \label{fig:dyck_scaling}
\end{figure}

\begin{figure}[ht]
  \centering
  \begin{subfigure}[b]{0.48\textwidth}
    \includegraphics[width=\textwidth]{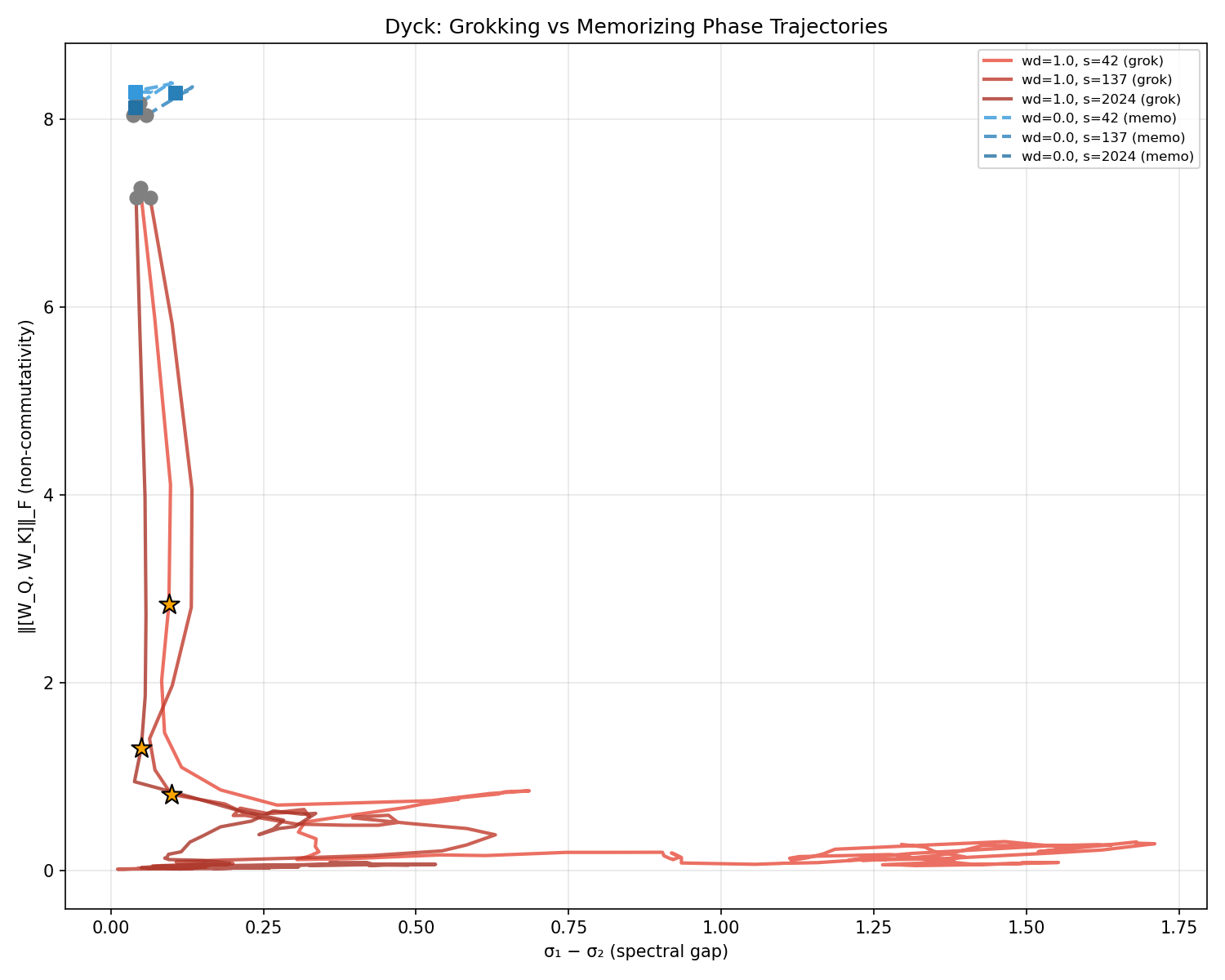}
    \caption{Dyck: Grok ($\lambda\!=\!1$) vs.\ control ($\lambda\!=\!0$)}
  \end{subfigure}
  \hfill
  \begin{subfigure}[b]{0.48\textwidth}
    \includegraphics[width=\textwidth]{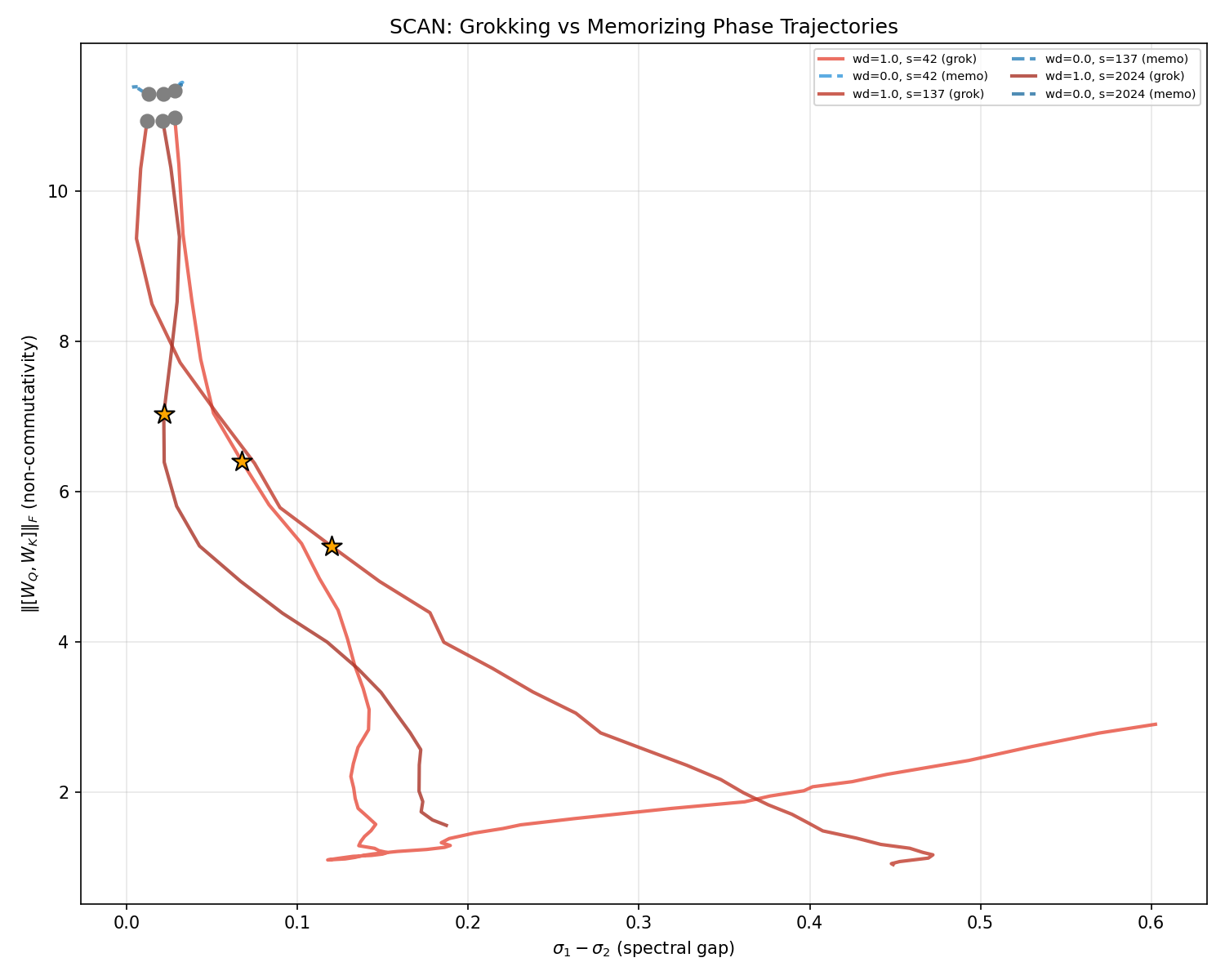}
    \caption{SCAN: Grok ($\lambda\!=\!1$) vs.\ control ($\lambda\!=\!0$)}
  \end{subfigure}
  \caption{\textbf{Grokking vs.\ memorizing in phase space.} Grokking runs (warm colors, 3 seeds) traverse the full phase space from competition (top-left) to alignment (bottom-right), while no-weight-decay controls (cool colors, 3 seeds) remain confined to a small region near initialization. Orange stars mark grokking events at the trajectory elbow. This topological distinction---sweep vs.\ stagnation---holds across all seeds and both datasets, with zero false positives.}
  \label{fig:spectral_grok_ctrl}
\end{figure}



\clearpage
\begin{thebibliography}{20}

\bibitem[Barak et al.(2022)]{barak2022hidden}
Barak, B., Edelman, B., Goel, S., Kakade, S., Malach, E., and Zhang, C.
\newblock Hidden progress in deep learning: SGD learns parities near the computational limit.
\newblock In \emph{NeurIPS}, 2022.

\bibitem[Lake and Baroni(2018)]{lake2018generalization}
Lake, B.~M. and Baroni, M.
\newblock Generalization without systematicity: On the compositional skills of sequence-to-sequence recurrent networks.
\newblock In \emph{ICML}, 2018.

\bibitem[Liu et al.(2022)]{liu2022omnigrok}
Liu, Z., Michaud, E.~J., and Tegmark, M.
\newblock Omnigrok: Grokking beyond algorithmic data.
\newblock In \emph{ICLR}, 2023.

\bibitem[Liu et al.(2023)]{liu2023grokking}
Liu, Z., Kitouni, O., Nolte, N., Michaud, E.~J., Tegmark, M., and Williams, M.
\newblock Towards understanding grokking: An effective theory of representation learning.
\newblock In \emph{NeurIPS}, 2023.

\bibitem[Loshchilov and Hutter(2019)]{loshchilov2019decoupled}
Loshchilov, I. and Hutter, F.
\newblock Decoupled weight decay regularization.
\newblock In \emph{ICLR}, 2019.

\bibitem[Lyu et al.(2023)]{lyu2023dichotomy}
Lyu, K., Jin, J., Li, Z., Du, S.~S., Lee, J.~D., and Hu, W.
\newblock Dichotomy of early and late phase implicit biases can provably induce grokking.
\newblock In \emph{ICLR}, 2024.

\bibitem[Murty et al.(2023)]{murty2023grokking}
Murty, S., Sharma, P., Andreas, J., and Manning, C.~D.
\newblock Grokking of hierarchical structure in vanilla transformers.
\newblock In \emph{ACL}, 2023.

\bibitem[Nanda et al.(2023)]{nanda2023progress}
Nanda, N., Chan, L., Lieberum, T., Smith, J., and Steinhardt, J.
\newblock Progress measures for grokking via mechanistic interpretability.
\newblock In \emph{ICLR}, 2023.

\bibitem[Power et al.(2022)]{power2022grokking}
Power, A., Burda, Y., Edwards, H., Babuschkin, I., and Misra, V.
\newblock Grokking: Generalization beyond overfitting on small algorithmic datasets.
\newblock In \emph{MATH-AI Workshop, ICLR}, 2022.

\bibitem[Xu(2026a)]{xu2026manifold}
Xu, Y.
\newblock Low-dimensional execution manifolds in transformer learning dynamics: Evidence from modular arithmetic tasks.
\newblock \emph{arXiv preprint arXiv:2602.10496}, 2026.

\bibitem[Xu(2026b)]{xu2026integrability}
Xu, Y.
\newblock Low-dimensional and transversely curved optimization dynamics in grokking.
\newblock \emph{arXiv preprint}, 2026.

\bibitem[Xu(2026a)]{xu2026spectral_edge}
Xu, Y.
\newblock The spectral edge thesis: A mathematical framework for intra-signal phase transitions in neural network training.
\newblock \emph{arXiv preprint arXiv:2603.28964}, 2026.

\bibitem[Yun et al.(2020)]{yun2020transformers}
Yun, C., Bhojanapalli, S., Rawat, A.~S., Reddi, S.~J., and Kumar, S.
\newblock Are transformers universal approximators of sequence-to-sequence functions?
\newblock In \emph{ICLR}, 2020.

\end{thebibliography}
\end{document}